\documentclass[runningheads]{llncs}
\usepackage{amsmath}

\usepackage{amsthm}

\usepackage[T1]{fontenc}
\usepackage{graphicx}
\usepackage{subcaption}
\usepackage{booktabs}

\usepackage{amsmath}
\usepackage{amssymb}
\usepackage{mathtools}
\usepackage{amsthm}
\usepackage{physics}

\newlength{\tempdima}
\newcommand{\rowname}[1]
{\rotatebox{90}{\makebox[\tempdima][c]{\textbf{#1}}}}

\begin{document}
\title{Liquid Resistance Liquid Capacitance Networks}
%
%
\author{M\'{o}nika Farsang\inst{1} \and
Sophie A. Neubauer\inst{2} \and
Radu Grosu\inst{1}}
\authorrunning{M. Farsang et al.}
%
\institute{Technische Universität Wien (TU Wien), Vienna, Austria\\
\email{\{monika.farsang, radu.grosu\}@tuwien.ac.at}\\
\and
DatenVorsprung GmbH, Vienna, Austria\\
\email{sophie@datenvorsprung.at}}
\maketitle              
\begin{abstract}
We introduce liquid-resistance liquid-capacitance neural networks (LRCs), a neural-ODE model which considerably improves the generalization, accuracy, and biological plausibility of electrical equivalent circuits (EECs), liquid time-constant networks (LTCs), and saturated liquid time-constant networks (STCs), respectively. We also introduce LRC units (LRCUs), as a very efficient and accurate gated RNN-model, which results from solving LRCs with an explicit Euler scheme using just one unfolding. We empirically show and formally prove that the liquid capacitance of LRCs considerably dampens the oscillations of LTCs and STCs, while at the same time dramatically increasing accuracy even for cheap solvers. We experimentally demonstrate that LRCs are a highly competitive alternative to popular neural ODEs and gated RNNs in terms of accuracy, efficiency, and interpretability, on classic time-series benchmarks and a complex autonomous-driving lane-keeping task. 

\end{abstract}

\section{Introduction}
\label{sec:intro}
Electrical equivalent circuits (EECs) are the foremost mathematical model used in neuroscience for capturing the dynamic behavior of biological neurons~\cite{kandel2000principles,wicks1996dynamic}. EECs associate the membrane of a neuron with a capacitor, whose potential varies according to the sum of the stimulus, leaking, and synaptic currents, passing through the membrane. Consequently, EECs are formulated as a set of capacitor ordinary differential equations (ODEs).

EECs, however, received little attention in the ML community due to their ODE nature, except for continuous-time recurrent neural networks (CT-RNNs) \cite{ct-rnns}, which are arguably EECs with electrical synapses~\cite{farsang2024learning}. This changed with the introduction of Neural ODEs~\cite{neuralODEs} and later their latent~\cite{latentODEs} and augmented~\cite{augmentedODEs} versions. EECs with chemical synapses are called liquid time-constant networks (LTCs)~\cite{Hasani2020LiquidTN}, as their time constant depends on both the state and the input.

The main promise of LTCs and Neural ODEs is to better capture physical reality and bridge the gap between natural sciences and ML. They are able to recover missing data in irregularly-sampled time series, which was investigated in LTCs~\cite{Hasani2020LiquidTN}, the GRU-ODE-Bayes method~\cite{gruODEbayes}, Neural Controlled Differential Equations~\cite{controlledNeuralODEs} and the Neural Lad framework \cite{neuralLAD}. Furthermore, LTCs were shown to be more interpretable in an autonomous lane-keeping task both in real-world~\cite{2020NeuralCP} and simulation environments~\cite{farsang2024learning}. An important obstacle in the use of Neural ODEs and especially LTCs, however, is their stiff oscillatory nature, which requires the use of expensive ODE solvers~\cite{neuralFlows}. 

In this paper, we show that the stiff dynamic behavior of LTCs results from abstraction. For simplicity, EECs and LTCs first ignore the saturation aspects of membrane's ionic channels~\cite{saturation}, as discussed in~\cite{farsang2024learning}. To take such aspects into account, they introduce  
saturated LTCs (called STCs). Second, and more importantly, EECs, LTCs and STCs also assume that membranal capacitance is constant, which is disproved by recent results demonstrating a nonlinear dependence on the neural state and input~\cite{capacitanceDepOnFrequency,capacitanceDepOnDaytime,capacitanceDepOnState}.

Here, we show that adding a liquid membrane capacitance to STCs, considerably reduces their variation with respect to inputs, while also enhancing their accuracy.
We call the resulting model a liquid-resistance liquid-capacitance neural network (LRC), as the LRC's liquid time-constant is now factored this way. Due to the gentle varying LRC behavior, an explicit Euler integration scheme with one unfolding, is often enough to obtain a very small error. We call LRCs solved with this scheme, LRC units (LRCUs). They closely resemble and improve on the performance of gated recurrent units, such as Long Short-Term Memory (LSTMs)~\cite{10.1162/neco.1997.9.8.1735}, Gated Recurrent Units (GRUs)~\cite{cho2014learning} and Minimal Gated Units (MGUs)~\cite{zhou2016minimal}. For irregularly sampled time series, however, one can use more expensive solvers with LRCs, too.

We experimentally evaluate LRCs with respect to their accuracy and speed on the classic gated-RNN and Neural-ODE benchmarks, respectively, and with respect to their interpretability on a complex autonomous-driving lane-keeping task. Our results show that LRCs outperform LSTMs, GRUs, and MGUs on RNN benchmarks, are able to successfully solve Neural-ODE tasks, and at the same time, are much more interpretable than LSTMs, GRUs, and MGUs on the lane-keeping task.

In summary, the main results of our paper are as follows:
\begin{itemize}
    \item We introduce Liquid-Resistance Liquid-Capacitance  Networks (LRCs),  which considerably improve the generalization, accuracy, and biological plausibility of EEC, LTC, and STC models, respectively.
    \item We introduce LRC Units (LRCUs), as a very efficient and accurate gated RNN-model, which results from solving LRCs with an explicit Euler scheme of order one and one unfolding, only.
    \item We prove that the liquid capacitance of LRCs, leads to a more gentle varying behavior than in LTCs and STCs, while at the same time dramatically increasing their accuracy even for cheap solvers.
    \item We experimentally show that LRCs are competitive in accuracy, efficiency, and interpretability to gated RNNs and Neural ODEs on classic benchmarks, and on a complex Lane-Keeping Task. 
\end{itemize}

The rest of this paper is organized as follows. In Section~\ref{sec:background}, we review EECs, LTCs, and STCs. Next, we introduce Liquid-Resistance Liquid-Capacitance Networks (LRCs) and LRCUs and discuss and prove their properties. In Section~\ref{sec:results} we provide our very encouraging experimental results. Finally, in Section~\ref{sec:conclusion}, we draw our conclusions and discuss future work. 

\section{Background}
\label{sec:background}

\subsection{Liquid Time-Constant Networks (LTCs)}
\label{sec:ltcs}

{\bf Definition.} EECs are a simplified electrical model, defining the dynamic behavior of the membrane-potential (MP) of a post-synaptic neuron, as a function of the MP of its pre-synaptic neurons, for electrical or chemical synapses~\cite{kandel2000principles,wicks1996dynamic}. EECs with chemical synapses only, are also known in the ML community as LTCs~\cite{worm_inspired,Hasani2020LiquidTN,2020NeuralCP}. An LTC with $m$ neurons and $n$ inputs is defined as follows:
\begin{equation}
\label{eq:ltcs}
\begin{array}{c}
    \dot{h}_i =-f_i\,h_i + u_i\,e_{li}\\[2mm]
    f_{i} = \sum_{j=1}^{m+n} g_{ji}\sigma(a_{ji}y_j+b_{ji}) + g_{li}\\[2mm]
    u_{i} = \sum_{j=1}^{m+n} k_{ji}\sigma(a_{ji}y_j+b_{ji}) + g_{li}.\\
\end{array}
\end{equation}

It states that the rate of change of the MP $h_i$ of neuron $i$, is the sum of its forget current $-f_i\,h_i$ and its update current $u_i\,e_{li}$. The forget conductance $f_i$ is the liquid time-constant reciprocal of $h_i$, which depends on $y\,{=}\,[h,x]$, where $h$ represents the MPs of presynaptic neurons and $x$ is the external inputs. Here, $g_{ji}$ is the maximum conductance of the synaptic channels, $a_{ji},b_{ji}$ are parameters controlling the sigmoidal probability of these channels to be open, and $g_{li}$, $e_{li}$ are the membrane's leaking (resting) conductance and potential, respectively.  In the update conductance $u_i$, parameters $k_{ji}\,{=}\,g_{ji}e_{ji}{/}e_{li}$ take the sign of $e_{ji}{/}e_{li}$, where $e_{ji}$ is the synaptic channel's reversal potential, that is, the MP at which there is no net ionic flow.

\subsection{Saturated Liquid Time-Constant Networks (STCs)}
\label{sec:sltc}

{\bf Definition.} For simplicity, LTCs ignore saturation in synaptic channels~\cite{saturation}. One can take this aspect into account, by constraining  the forget and signed-update conductances $f_i$ and $u_i$ with a sigmoid and a hyperbolic tangent, to range within [0,1] and [-1,1], respectively. Letting $f_i$ and $u_i$ be as above, this leads to the definition of saturated LTCs (STCs) of~\cite{farsang2024learning}: 
\begin{equation}
\label{eq:sltcs}
\begin{array}{c}
    \dot{h}_i =-\sigma(f_i)\,h_i + \tau(u_i)\,e_{li}.
\end{array}
\end{equation}

\section{Liquid-Resistance Liquid-Capacitance Networks (LRCs)}
\label{sec:LRCs}
\subsection{Model Definition}

{\bf Definition.} LTCs and STCs assume that the membrane capacitance is constant and equal to one. This is however not the case in biological neurons. They have a nonlinear dependence on the MPs and the incoming inputs~\cite{capacitanceDepOnFrequency,capacitanceDepOnDaytime,capacitanceDepOnState}. LRCs take this aspect into account, by adding a liquid elastance (reciprocal of capacitance) $\epsilon(w_i)$, in Equation~\eqref{eq:sltcs}, where the weights matrix $o$ and the bias vector $p$ play the same role as the ones in an artificial neuron:
\begin{equation}
\label{eq:lcrs}
\begin{array}{c}
    \dot{h}_i =\epsilon(w_i)\,(-\sigma(f_i)\,h_i + \tau(u_i)\,e_{li})\\[2mm]
    w_{i} = \sum_{j=1}^{m+n} o_{ji}y_j+ p_{i}.
\end{array}
\end{equation}
Thus, the time constant in LRCs factors into a liquid-resistance and a liquid-capacitance. The second, acts as an additional control of the time constant, by increasing it when the desired variation of the function is gentle, and leaving it unchanged when the desired behavior is stiff. For convenience, we provide the definition of both an asymmetric and a symmetric form of elastance in Equations~(\ref{eq:asymmetrice}-\ref{eq:symmetrice}), respectively. Here, the parameters vector $k$ of the symmetric version is non-negative in each of its components:
%
%
\begin{align}
    (A) \quad &\epsilon(w_i) = \sigma (w_i), \label{eq:asymmetrice}\\
    (S) \quad &\epsilon(w_i) = \sigma (w_{i} + k_i) - \sigma (w_{i} - k_i). \label{eq:symmetrice}
\end{align}

\begin{figure}[t]
  \centering
  \includegraphics[width=0.45\linewidth]{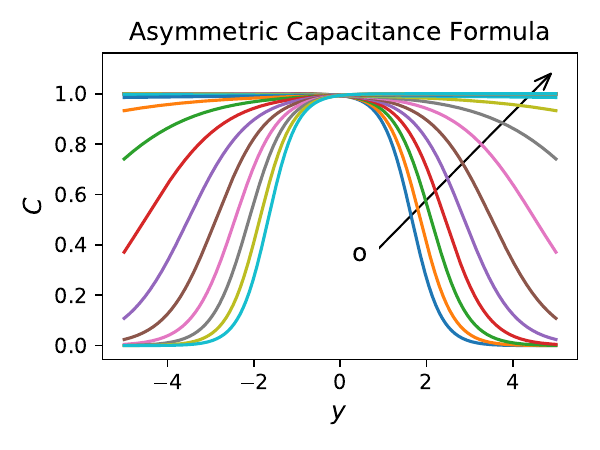}
  \includegraphics[width=0.45\linewidth]{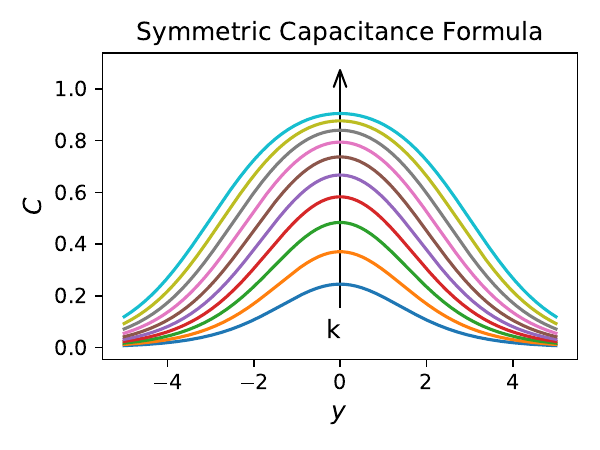}
  \caption{The asymmetric and symmetric shape of the membrane's elastance $\epsilon(w_i)$ proposed for LRCs, using Formulas~\eqref{eq:asymmetrice} and \eqref{eq:symmetrice}, respectively. 
  }
  \vspace*{-3mm}
  \label{fig:elastances}
\end{figure} 

%

\begin{figure}[tb]
  \centering
  \includegraphics[height=0.275\linewidth]{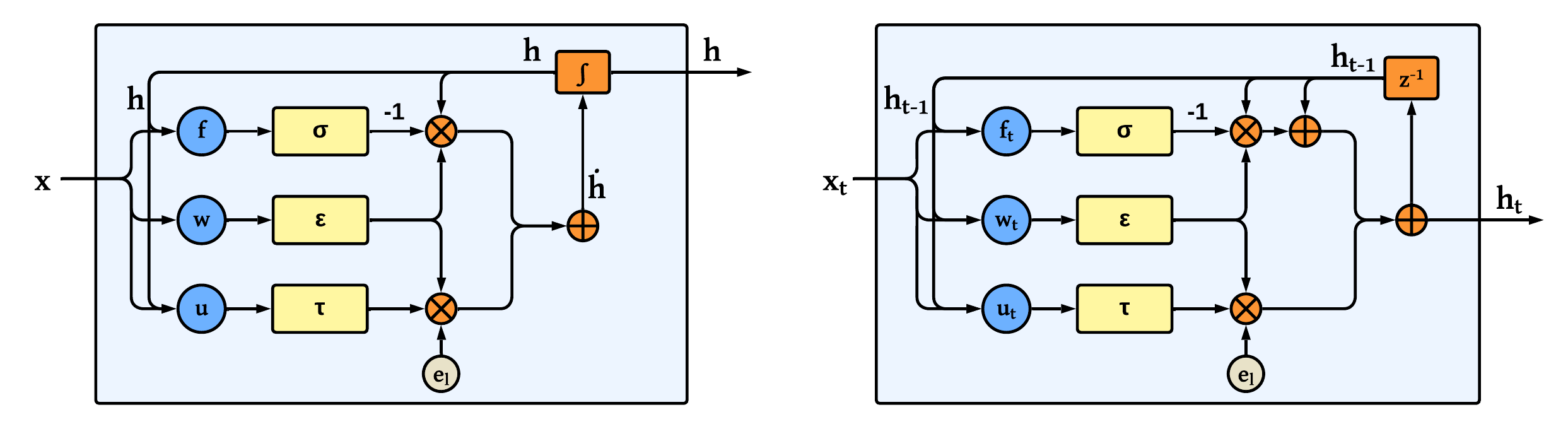}
  \caption{The architecture of LRCs (left) and of LRCUs (right, explicit 1st order Euler discretization of LRCs with one unfolding). Both LRCs and LRCUs have a sigmoidal forget gate $\sigma(f_t)$, a hyperbolic-tangent update gate $\tau(u_t)$, and a sigmoidal symmetric or asymmetric smoothen gate $\epsilon(w_t)$. }
  \label{fig:LRCs}
\end{figure}

The shape of the two elastances is shown in Figure~\ref{fig:elastances}. The first is asymmetric with respect to $y$, while the second is symmetric with respect to $y$. They both scale the elastance to range within [0,1], thus acting to possibly decrease the variation of $h$. The linear output layer of the LRC may compensate this, by possibly having values larger than one. The architecture of LRCs is given in Figure~\ref{fig:LRCs} (left).

%
%

\subsection{Model Properties}
In this section, we investigate important model properties of LRCs. For this purpose, we focus on the complex autonomous driving Lane-Keeping experiment, as it is a regression task where LTCs have been shown to perform well. We aim to determine if we can improve upon the properties of LTCs by incorporating the additional saturation aspect and liquid capacitance of LRCs. Specifically, we are interested in examining the Lipschitz constant, model accuracy, and efficiency.

\begin{figure}[t]
\centering
\begin{subfigure}{.33\linewidth}
  \centering
  \includegraphics[width=\linewidth]{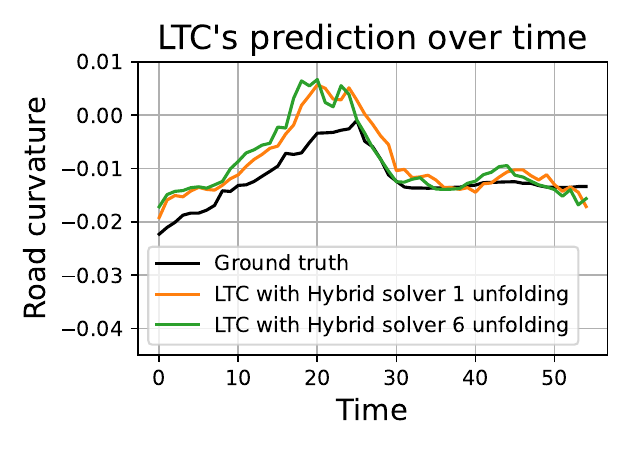}
  \label{fig:ltc_unfoldings}
\end{subfigure}%
\begin{subfigure}{.33\linewidth}
  \centering
  \includegraphics[width=\linewidth]{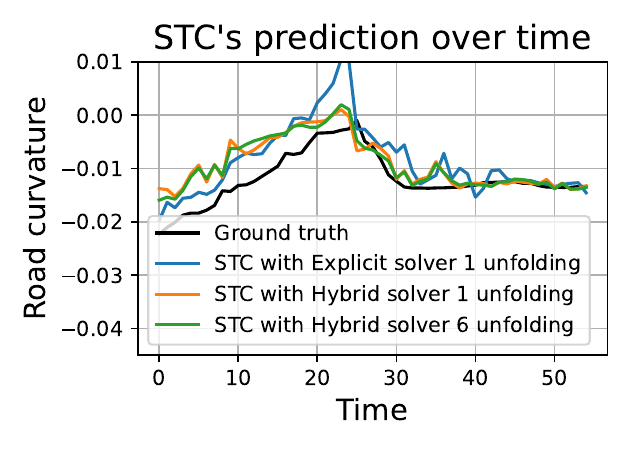}
  \label{fig:sltc_unfoldings}
\end{subfigure}%
\begin{subfigure}{.33\linewidth}
  \centering
  \includegraphics[width=\linewidth]{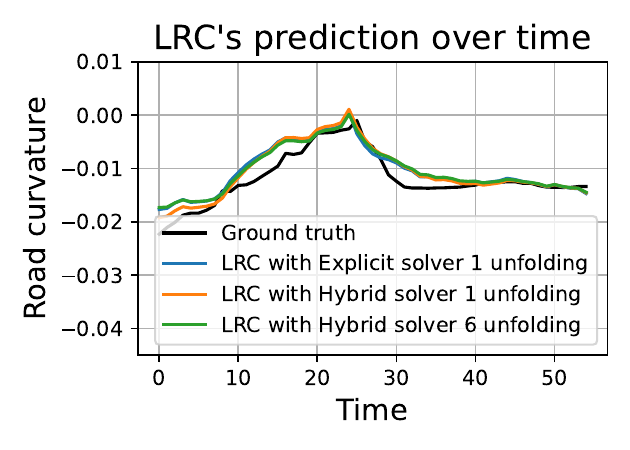}
  \label{fig:gcu_unfoldings}
\end{subfigure}
\caption{Dynamic behavior of the output of LTC, STC and LRC in the regression task of Lane-Keeping~\cite{lechner2022all}. 
}
\label{fig:behavior}
\end{figure}

\begin{figure}[tb]
  \centering
  \includegraphics[width=0.75\linewidth]{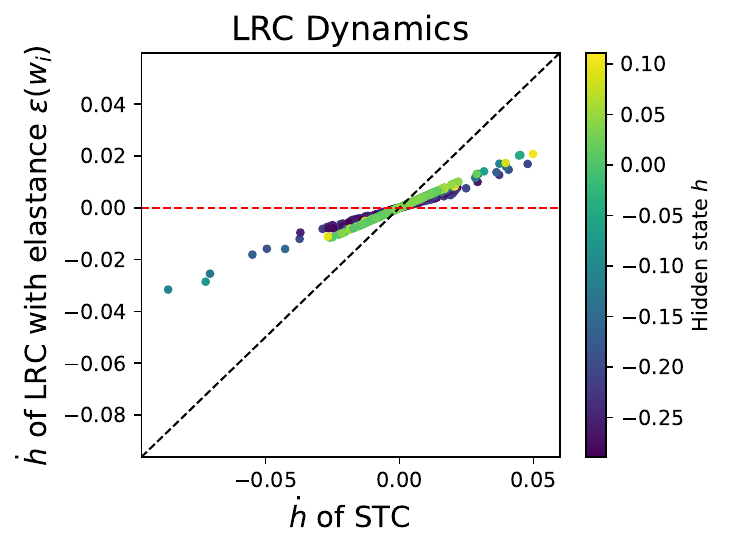}
  \caption{Experiment showing that the LRC's elastance dampens the dynamic behavior of STCs in an input and state-dependent way. It can fully reduce it to zero, or keep it as it is, indicated by the red and black dashed lines, respectively. 
  }
  \vspace*{-3mm}
  \label{fig:dynamicsLRC}
\end{figure}

\paragraph{Lipschitz constant.} In the case of LTCs, shown on the left in Figure~\ref{fig:behavior}, the output has an oscillatory behavior, with a relatively large validation loss. Moreover, LTCs fail to converge with an explicit Euler integration scheme of order one and just one unfolding. In contrast, STCs convergence with an explicit Euler integration scheme of order one and just one unfolding presented in the middle figure. However, the output neuron still exhibits stiff oscillatory behavior. In LRCs, this neuron has a very gentle varying behavior, with a very small validation loss, even when using an explicit Euler scheme of order one and just one unfolding (no overhead). This highly advantageous, as it was demonstrated that a lower Lipschitz constant enhances robustness~\cite{cisse2017parseval}, generalization~\cite{sokolic2017robust}, and interpretability~\cite{tsipras2018there}.




Due to its very gentle varying behavior of LRCs, an explicit Euler integration with one unfolding, is often sufficient for obtaining an acceptable validation loss. This dramatically speeds up computation, as the solver introduces no overhead. This is particularly useful for time series with regularly sampled data, lacking timing information. Here, a time step of one, leads to a network closely resembling gated recurrent neural networks~\cite{10.1162/neco.1997.9.8.1735,cho2014learning,zhou2016minimal}. Its architecture is shown in Figure~\ref{fig:LRCs} (right). We call this network, an LRC unit (LRCU). For further details and discussion on this, see Appendix~\ref{appendix:lrcu_as_gatedrnn}. As we show in the experimental-results section in Section~\ref{sec:er:rnns}, LRCUs outperform gated RNNs on the RNN benchmarks, when used wih either a symmetrical elastance (LRCU-S) or an asymmetrical elastance (LRCU-A).

\begin{theorem}[LRC Lipschitz]\label{thm:smoothness}
    Let the Lipschitz constant $\lambda^s$ bound the sensitivity of an STC-model instance of Equation~\eqref{eq:sltcs}, with respect to its inputs and hidden states. Then, there exists an associated LRC-model instance of Equation~\eqref{eq:lcrs}, with a Lipschitz constant $\lambda^r$, such that $\lambda^r<\lambda^s$.
 \end{theorem}
The proof of Theorem~\ref{thm:smoothness} is given in the Appendix. Intuitively, for an LRC having the parameter values for its forget and update conductances fixed as in an STC, one can set the parameter values of its elastance, such that the Lipschitz constant of the LRC is smaller than the one of the STC. This leads to LRC being a function which does not change rapidly when the input changes. 

\begin{figure}[t]
\centering
\begin{subfigure}[!b]{.49\linewidth}
  \centering
  \includegraphics[width=\linewidth]{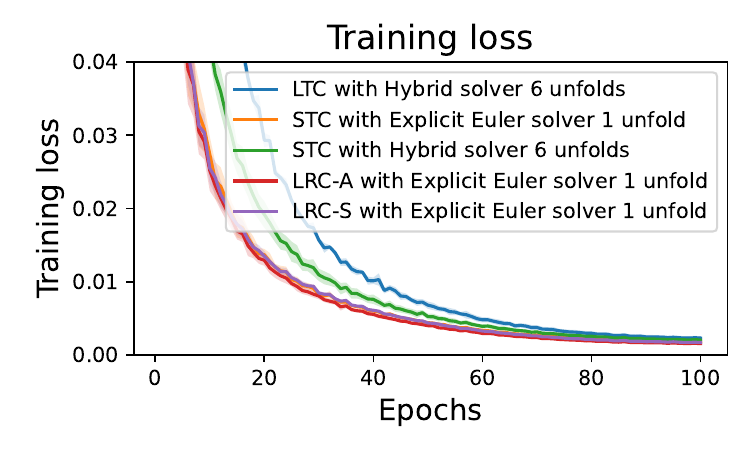}
  \label{fig:sltc_training_loss}
\end{subfigure}%
\begin{subfigure}[!b]{.49\linewidth}
  \centering
  \includegraphics[width=\linewidth]{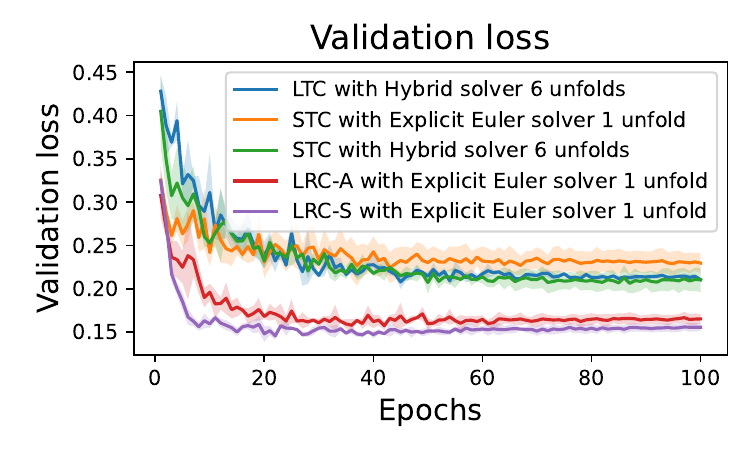}
  \label{fig:validation_loss}
\end{subfigure}
\caption{Training loss in the first and validation loss in the second figure of LTCs, STCs, and LRCs for the Lane-Keeping Task.} 
\label{fig:loss}
\end{figure}

\paragraph{Accuracy and Generalization.} The addition of a liquid elastance in LRCs does not only reduce oscillations, but it also dramatically improves accuracy. This is illustrated for the Lane-Keeping Task in Figure~\ref{fig:loss}, where the validation loss of the LRC's output neuron, is considerably smaller compared to the one of LTCs and STCs. While the above theorem proves a smaller Lipschitz constant, the following theorem proves generalization.

\begin{theorem}[LRC Generalization]\label{thm:accuracy}
    Let $h^s$ be an STC-model instance of Equation~\eqref{eq:sltcs} with $h^r$ being an associated LRC-model instance of Equation~\eqref{eq:lcrs} with $\lambda^r < \lambda^s$ (Theorem~\ref{thm:smoothness}). Let $o^T_t$ be training labels and $\hat o^{T,r}_t$  and $\hat o^{T,s}_t$ be the predictions of the models $h^r$ and $h^s$ respectively. Let both models have a small training loss $\|o^T_t-\hat{o}^{T}_t\|_2 < \epsilon_T$ with small $\epsilon_T>0$. Let $y^T_t$ and $y^V_t$ be respectively training and validation input. Let the validation set be similar to the training set (drawn in-distribution) in a sense that for every time $t$ it holds that $\|y^T_t-y^V_t\|_2<\epsilon_y$ and $\|o^T_t-o^V_t\|_2<\epsilon_o$ with small $\epsilon_y, \epsilon_o > 0$. Then it holds that the upper bound of the validation loss of the LRC-model is smaller than the one of the STC-model.
 \end{theorem}
The proof of Theorem~\ref{thm:accuracy} is given in the Appendix. Intuitively, if the validation set is similar to the training set modulo some small perturbation, are drawn from the same distribution, one can use a Taylor series approximation of the validation loss $o(y^V) \approx o(y^T) + (y^V-y^T)^T\nabla_y o(y^V)$ and get an upper bound on the validation loss by using Theorem~\ref{thm:smoothness}. As the Lipschitz constant of the LRC is smaller than the one of STC, also the upper bound of the validation loss of the LRC-model is smaller than the one of the STC-model.


\paragraph{Efficiency.} Solving LTCs, STCs, and LRCs with popular ODE-integration techniques, is computationally expensive. A hybrid Euler method~\cite{2020NeuralCP} for example, takes multiple unfolding rounds for a given input, to achieve a good approximation of the solution. In Figure~\ref{fig:loss} we show the validation loss of LTCs, STCs, and LRCs, for the Lane-Keeping Task~\cite{lechner2022all}, where the state of the neurons is computed with a fixed integration step $\Delta$. Diving this into 6 equidistant time steps $\Delta{/}6$ within one update step, results in a considerably better model, but at a higher cost of 10min versus 4min per epoch. This results in a $2.5\times$ longer computation time. 

\begin{table}[]
    \centering
    \caption{Training details of LTC, STCs and LRCs on the Lane Keeping dataset.}
    \begin{tabular}{lr@{\hskip 0.1in}r@{\hskip 0.1in}r}
          \toprule
          Model &  Time Per Epoch & Best Epoch & Lipschitz Constant\\
          \midrule
          LTC H6    & 10.06 min & $52 \pm 06$ & $90{\pm}23{\cdot}10^{-4}$\\
          STC E1    & 03.97 min & $50 \pm 22$ & $101{\pm}25{\cdot}10^{-4}$\\
          STC H6    & 10.13 min & $69 \pm 10$ & $89{\pm}23{\cdot}10^{-4}$\\
          LRC-A E1  & 04.05 min & $43 \pm 08$ & $47{\pm}08{\cdot}10^{-4}$\\
          LRC-S E1  & 04.16 min & $22 \pm 03$ & $43{\pm}11{\cdot}10^{-4}$\\
          \bottomrule
    \end{tabular}
    \label{tab:my_label}
\end{table}

Table~\ref{tab:my_label} presents that per epoch and curves, the training took around 10 minutes for 6 unfoldings per step and 4 minutes for 1 unfolding per step, which is 2.5 times faster. As the curves of LTCs and STCs with the hybrid solver converge in around 50-70 epochs, and the LRC-S and LRC-A in around 20 and 40, respectively, the total speedup until the best epoch of the latter with respect to the former is 3-7.8 times, while also being a much more accurate. Note that solving STCs with Explicit Euler and one unfolding is computationally inexpensive, but its validation loss is not very satisfactory.

In Figure~\ref{fig:loss}, we also compare the training and validation loss of the LRCUs and LRCs, for the Lane-Keeping Task, where we compute the elastance term $\epsilon(w_i)$ in two different ways, by using Equations~(\ref{eq:asymmetrice}-\ref{eq:symmetrice}), and learning the parameters $o$, $p$, and $k$. Using these proposed liquid elastances leads to a considerably faster convergence of validation loss, to a much smaller loss. In particular, the symmetric elastance achieves the best results, by essentially converging in around 20 epochs, to a loss of about 0.15. The LTCs and STCs instead, converge in about 60 epochs, at a loss of 0.22.

\section{Further Experimental Results}\label{sec:results}

In the previous sections we defined LRCs as an extension of STCs, by arguing that using a constant capacitance, as is the case in LTCs and STCs, is not only biologically implausible, but it also leads to large oscillations in the trained networks. We also showed that by considering a liquid (that is a state-and-input dependent) capacitance, dramatically reduces the validation loss of the resulting network, by having better generalization. Due to the more gently varying behavior, solving LRCs with explicit, one-unfolding Euler, leads to very small validation loss, without any computational overhead. We called LRCs integrated this way, LRCUs, as they closely resemble gated RNNs when using a time-step of one. 

In the following subsections, we present our further experimental results for LRCs, on three different topics of interest in the ML community. The first investigates LRCUs in comparison with the one of LSTMs, GRUs, and MGUs on classic benchmarks used for RNNs. The second investigates LRCs in solving classic Neural-ODE tasks. Finally, the third investigates the interpretability of the learned LRCUs (and thus LRCs) for the Lane-Keeping Task.

\subsection{LRCUs on RNNs Benchmarks}
\label{sec:er:rnns}
In this section we ask and positively answer, if LRCUs are competitive with respect to accuracy and convergence speed, compared to the popular gated RNNs. To this end, we conduct experiments on a wide range of time-series modeling applications, including: Classification of activities based on irregularly sampled localization data; IMDB movie reviews; and Permuted sequential MNIST tasks. 

To achieve a somewhat similar number of parameters, we use 64 cells for LRCUs, and 100 cells for the others as internal representation. The total number of parameters in a model also depends on the number of inputs and outputs of the tasks considered. Finally, we also conduct experiments in a relatively complex and high-dimensional image-based regression task, for lane keeping in autonomous vehicles. 
All experimental details, including the hyperparameters and the number of trainable parameters, are given in Appendix~\ref{appendix:experiment_details}. In Table~\ref{tab:accuracy} we provide the accuracy of the various models compared in the classification tasks. 

\begin{table}[t]
    \centering
    \caption{Accuracy in the classification datasets in percentage.} 
    \small
    \begin{tabular}{lccc}
        \toprule
        Model & Localization & IMDB & psMNIST\\
        \midrule
        LSTM & $82.90 \pm 0.31$ & $86.56 \pm 0.49$ & $91.20 \pm 0.10$\\
        GRU & $82.76 \pm 0.41$  & $86.32 \pm 0.51$ & $90.20 \pm 0.44$\\
        MGU & $83.35 \pm 0.30$ & $85.18 \pm 0.85$ & $87.78 \pm 0.82$\\
        LRCU-S  & $\mathbf{84.21\pm 0.26}$ & $85.73 \pm 0.41$ & $\mathbf{91.74 \pm 0.41}$\\ 
        LRCU-A  & $83.90 \pm 0.34$ & $\mathbf{87.00 \pm 0.53}$ & $91.31 \pm 0.33$\\ 
        \bottomrule        
    \end{tabular}
    \label{tab:accuracy}
\end{table}

\paragraph{Localization Data for Person Activity.}
The Localization Data for the Person Activity dataset given in~\cite{misc_localization_data_for_person_activity_196}, captures the recordings of five individuals, engaging in various activities. Each wore four sensors at the left and the right ankle, at the chest, and at the belt, while repeating the same activity five times.
The associated task, is to classify their activity based on the irregularly sampled time-series. This task is a classification problem, adapted from~\cite{lechner2020learning}. 


In Table \ref{tab:accuracy}, we present the experimental results for the accuracy of the classification for LSTMs, GRUs, MGUs, and LRCUs. The experiments for the LRCUs were done with both the asymmetric (LRCU-A) and the symmetric (LRCU-S) elastance, respectively. As one can see, both LRCUs considerably outperformed the traditional gated RNNs. In particular, the LRCU-S performed best, achieving an accuracy of 84.99\%, slightly better than the LRCU-A. 


\paragraph{IMDB Movie Sentiment Classification.}
The IMDB Movie-Review Sentiment Classification dataset, also known as the Large Movie Review Dataset~\cite{maas-etal-2011-learning}, is designed for binary sentiment classification. This data set includes 25,000 movie reviews, for both training and testing. Each review was labeled with either a positive or a negative sentiment. 


Table \ref{tab:accuracy} presents our experimental results for this task, too.
While the LRCU-S achieved a performance which was comparable to the one of the traditional RNNs, the LRCU-A had the best performance, by achieving an accuracy of 87\%. 

\paragraph{Permuted Sequential MNIST.}
The Permuted Sequential MNIST dataset is a variant of the MNIST Digits Classification data set, designed to evaluate recurrent neural networks, adapted from~\cite{le2015simple}. In this task, the 784 pixels (originally images of size $28{\times}28$) of digits are presented sequentially to the network. The main challenge lies in predicting the digit category, only after all pixels are observed. This task tests the network's ability to handle long-range dependencies. To make the task even more complex, a fixed random permutation of the pixels is first applied. 


Our experimental results are shown in Table~\ref{tab:accuracy}.
As before, both LRCUs surpass the accuracy of the other models. In particular, the LRCU-S achieves the best results.  

\begin{figure}[t]
  \centering
  \begin{subfigure}{0.48\linewidth}
    \centering
    \includegraphics[height=0.5\linewidth]{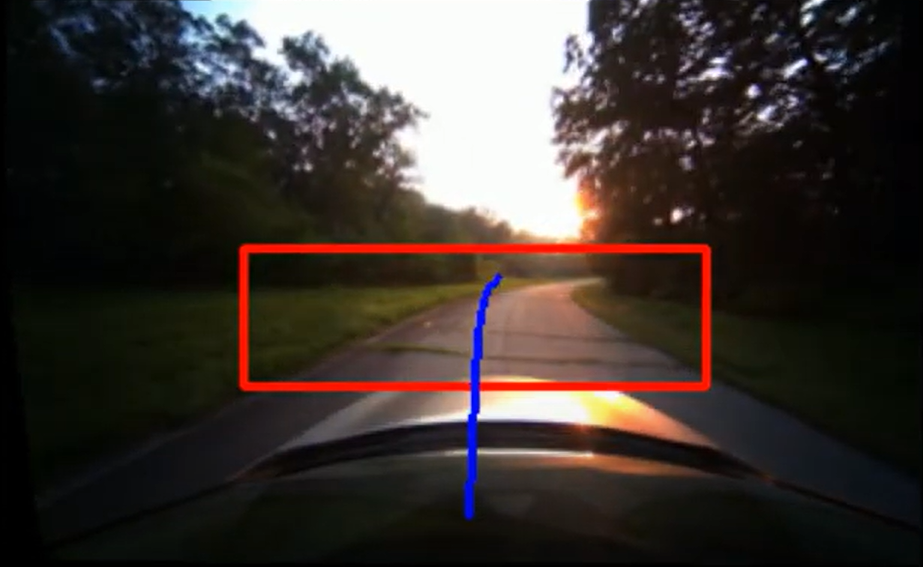}
  \end{subfigure}
  \begin{subfigure}{0.48\linewidth}
    \centering
    \includegraphics[height=0.5\linewidth]{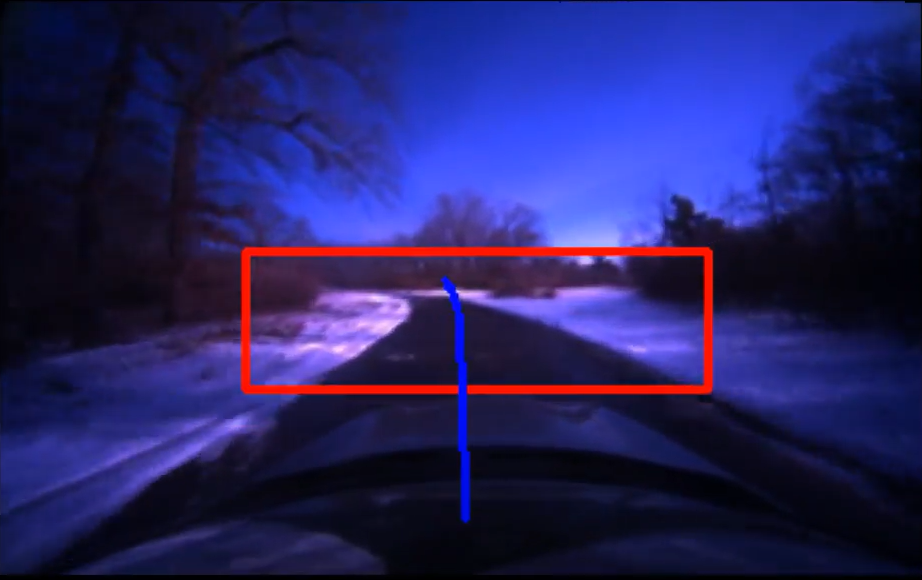}
  \end{subfigure}
  \caption{Lane-Keeping task. The first two figures show summer and winter conditions, respectively. The red rectangle indicates the input and the blue line the predicted steering angle of the network.}
  \label{fig:lane_keeping_images}
  \vspace{-2mm}
\end{figure} 
\begin{figure}[t]
  \centering
  \begin{subfigure}{0.6\linewidth}
    \centering
    \includegraphics[height=0.55\linewidth]{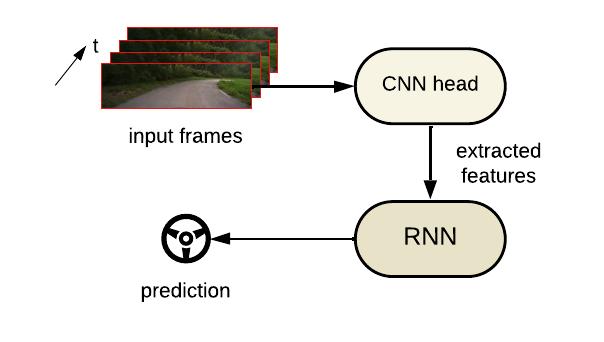}
  \end{subfigure}
  \caption{Network architecture for the Lane-Keeping task. The CNN head extracts the input features of a video stream. They are passed to the recurrent policy, responsible for steering.}
  \label{fig:lane_keeping_network}
  \vspace{-2mm}
\end{figure} 

\paragraph{Lane-Keeping Task.}
In the Lane-Keeping Task, the control agent is provided with the front-camera input, consisting of RGB images of size $48{\times}160$ pixels. It is required to autonomously navigate and maintain its position within the road, by properly predicting its curvature. The predicted road curvature corresponds to the steering action necessary for lane-keeping, and holds the advantage of being vehicle-independent, as the actual steering angle depends on the type of car used. 

The dataset for the Lane-Keeping Task task is obtained from human-driving recordings captured under various weather and illumination conditions~\cite{lechner2022all}, as shown in Figure \ref{fig:lane_keeping_images}. The network architecture illustrated in Figure~\ref{fig:lane_keeping_network} contains a CNN-head for extracting the main features from the camera input. They are fed into the gated recurrent models, for the sequential-regression prediction. This setup is adapted from~\cite{farsang2024learning}.

In Table~\ref{tab:lanekeeping_table}, we report the experimental losses of the models considered on the Lane-Keeping task. As it was the case before, both LRCU-A and LRCU-S obtain comparable validation losses, with LRCU-A achieving the best weighted-validation loss (loss weighted by road curvature). This is arguably the more important loss. 

We test the models in the closed-loop setting to evaluate their performance of robust lane keeping. By injecting extra Gaussian noise with variance 0.1 into the input images, we find that LRCUs can handle it without any crashes, and show better performance than other models in even higher Gaussian noise with variance 0.2 too, as reported in Table~\ref{tab:lanekeeping_crash_table}.

\begin{table}[t]
    \centering
    \caption{Mean squared loss results for the Lane-Keeping task, averaged over 3 seeds.}
    \begin{tabular}{lcc}
      \toprule
      Model & Validation Loss & W-Validation Loss \\
      \midrule
      LSTM & $\mathbf{0.139 \pm 0.008}$ & $0.013 \pm 0.0010$\\
      GRU & $0.145 \pm 0.006$ & $0.013 \pm 0.0010$ \\
      MGU & $\mathbf{0.139 \pm 0.007}$ & $0.012 \pm 0.0010$\\
      LRCU-A & $0.154 \pm 0.005$ & $\mathbf{0.010 \pm 0.0003}$\\ 
      LRCU-S & $0.142 \pm 0.004$ & $\mathbf{0.011 \pm 0.0010}$\\ 
      \bottomrule
    \end{tabular}
    \label{tab:lanekeeping_table}
\end{table}

\begin{table}[t]
    \centering
    \caption{Crash likelihood in closed-loop simulation with additional Gaussian noise in the Lane-Keeping task, averaged over 3 seeds. Values closer to zero indicate reliable, crash-free behavior.}
    \begin{tabular}{lcc}
      \toprule
      Model & Noise $\sigma^2=0.1$ & Noise $\sigma^2=0.2$ \\
      \midrule
      LSTM & $\mathbf{0.000 \pm 0.000}$ & $0.483 \pm 0.254$\\
      GRU & $0.133 \pm 0.189$ & $0.767 \pm 0.047$\\
      MGU & $0.267 \pm 0.221$ & $0.817 \pm 0.037$\\
      LRCU-A & $\mathbf{0.000 \pm 0.000}$ & $\mathbf{0.200 \pm 0.231}$\\ 
      LRCU-S & $\mathbf{0.000 \pm 0.000}$ & $0.300 \pm 0.216$\\ 
      \bottomrule
    \end{tabular}
    \label{tab:lanekeeping_crash_table}
\end{table}
\vspace*{-2ex}

\subsection{LRCs on Neural ODEs Benchmarks}
\label{sec:er:NeuralODEs}

LRCs are competitive in accuracy and convergence speed on popular Neural-ODE benchmarks. As examples, we consider a Periodic Sinusoidal, a Spiral towards the origin \cite{Chen2018NeuralOD}, the non-linear Duffing oscillator~\cite{riyazialeveraging,sholokhov2023physics,constanteamores2024enhancingpredictivecapabilitiesdatadriven} and three versions of the Lotka-Volterra (LV) predator-prey model~\cite{BHATTACHARYA201031,essay87568} well-known from evolutionary biology and ecology. 

In our experiments we use an LRC with 16 hidden states for internal representation. LRCs are highly suitable for such tasks as they are ODEs. We also use two extra layers for mapping. The former maps the state of these systems to the dimensionality of the hidden state, allowing us to learn more expressive 16-dimensional dynamics. The latter remaps these dynamics back to the original system state dimensionality. 

We compare the performance of the LRCs with a 3-layer Neural-ODE, containing 32, 32, and 2 neurons, respectively. These three layers match the number of trainable parameters, with the ones of the LRCs. Neural-ODEs are solved by Runge-Kutta 4(5) of Dormand-Prince, and LRCs by explicit Euler of order 1. The results of our experiments averaged over 3 seeds, are shown in Table~\ref{tab:ode_results}. 
\vspace*{-1ex}
\begin{table}[!h]
    \centering
    \caption{Test loss of various ODE tasks, with LRCs showing good performance. Results are averaged over 3 seeds.}
    \begin{tabular}{lcc}
      Task & Neural-ODE & LRC \\
      \midrule
      Sinusoid & $0.140 \pm 0.033$ &  $\mathbf{0.019 \pm 0.005}$\\ 
      Spiral & $0.012 \pm 0.005$ & $\mathbf{0.009 \pm 0.004}$\\ 
      Duffing Oscill. & $0.142 \pm 0.117$ & $\mathbf{0.003 \pm 0.001}$\\
      Periodic LV & $0.025 \pm 0.006$ & $\mathbf{0.005 \pm 0.001}$\\
      Asymptotic LV & $0.030 \pm 0.0002$ & $\mathbf{0.009 \pm 0.001}$\\
      Non-linear LV & $0.033 \pm 0.010$ & $\mathbf{0.008 \pm 0.004}$\\
      \bottomrule
    \end{tabular}
    \label{tab:ode_results}
\end{table}
\vspace*{-4ex}


\subsection{Interpretability of LRCUs on the Autonomous-Driving Lane-Keeping Task}
\label{sec:er:interpretability}

To show that LRCUs outperform LSTMs, GRUs, and MGUs in terms of interpretability, we conduct Lane-Keeping Task experiments, and compute the following interpretability metrics: the network attention and its robustness by the structural similarity index, the absolute correlation of neural activity with road trajectory, and the activity of neurons.

Understanding neural-network attention during decision-making is a crucial element in increasing trust in the network. By using the VisualBackprop~\cite{Bojarski2016VisualBackPropVC} method, we visualize which pixels in the input image have the most impact at the given timestep. Even though the same structure of CNN heads is used in the training pipeline, the different recurrent parts of the decision-making policy, have a big influence on the learned features in the convolutional part through backpropagation. The attention in summer and winter season is given in the Appendix. 

\begin{figure}[t!]
  \centering
  \includegraphics[width=0.48\linewidth]{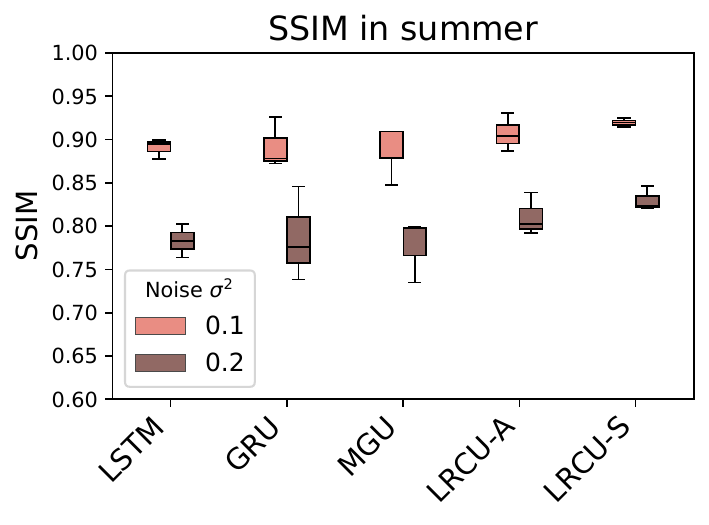}
  \includegraphics[width=0.48\linewidth]{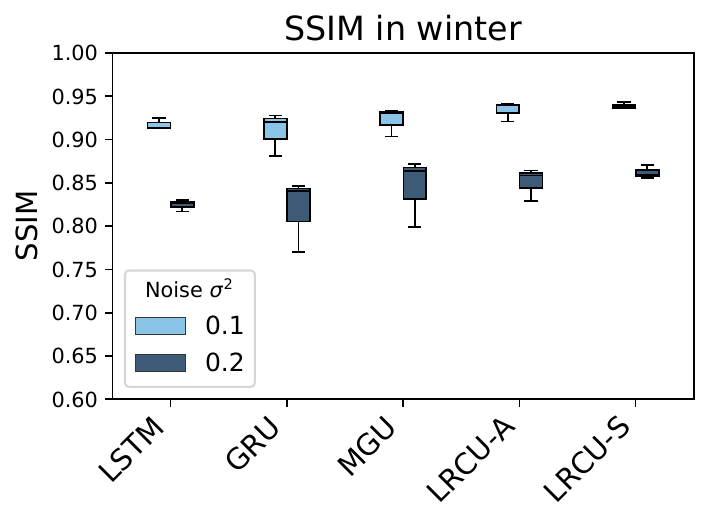} 
  \caption{Robustness of the attention, measured by the Structural Similarity Index (SSIM) \cite{SSIM} of the models, in summer (left) and in winter (right).}
  \label{fig:ssim}
  \vspace{-2ex}
\end{figure}

By injecting a zero mean Gaussian noise with a variance of 0.1 and 0.2 into the tests, respectively, we can measure how robust their attention is. Quantitatively, this is measured by the Structural Similarity Index, displayed in Figure~\ref{fig:ssim}. The LRCU models maintain the most similar focus of their attention in the presence of noise, indicated by the boxplots closer to 1. Lighter color refers to additional Gaussian noise of 0-mean and $\sigma^2=0.1$ variance, and darker to $\sigma^2=0.2$ variance.

\begin{figure}
\settoheight{\tempdima}{\includegraphics[width=.2\linewidth]{example-image-a}}%
\centering\begin{tabular}{@{}c@{ }c@{ }c@{ }c@{}c@{}c@{}}
& LSTM & GRU & MGU & LRCU-A & LRCU-S \\
\rowname{Cell 1 }&
\includegraphics[width=.19\linewidth]{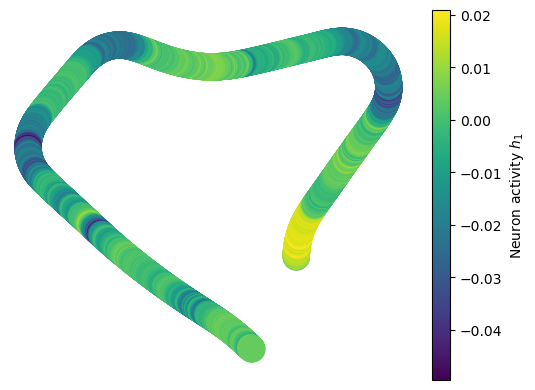}&
\includegraphics[width=.19\linewidth]{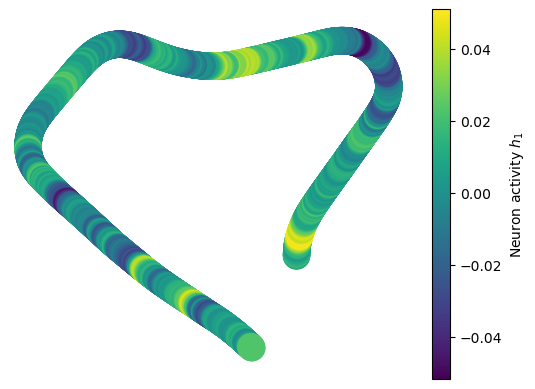}&
\includegraphics[width=.19\linewidth]{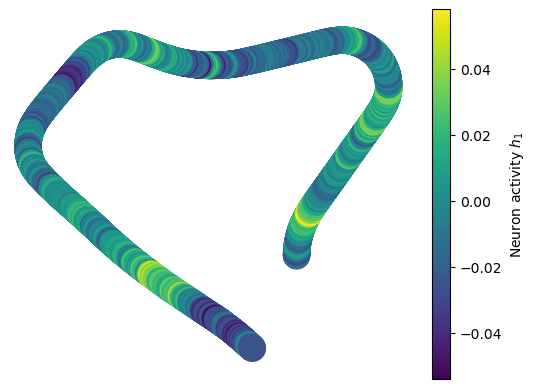}&
\includegraphics[width=.19\linewidth]{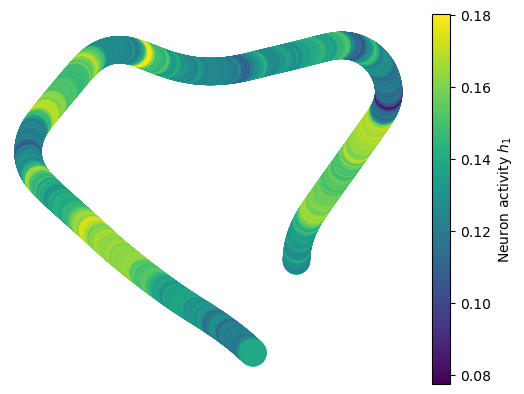}&
\includegraphics[width=.19\linewidth]{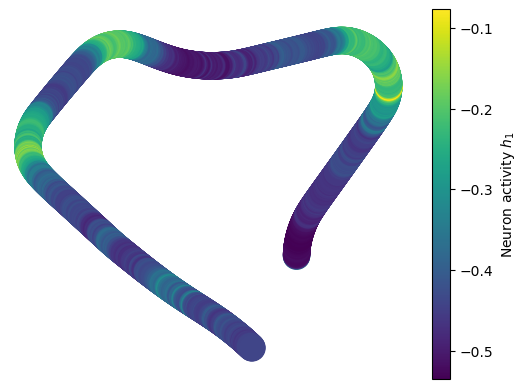}\\
\rowname{Cell 2}&
\includegraphics[width=.19\linewidth]{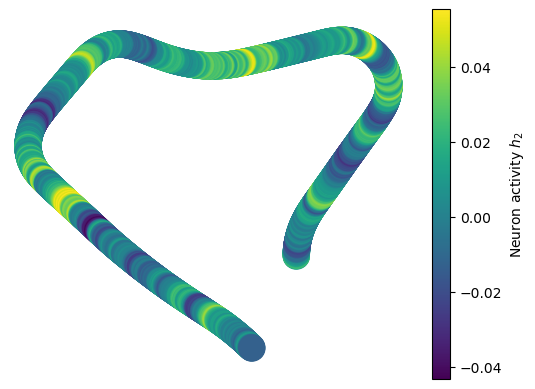}&
\includegraphics[width=.19\linewidth]{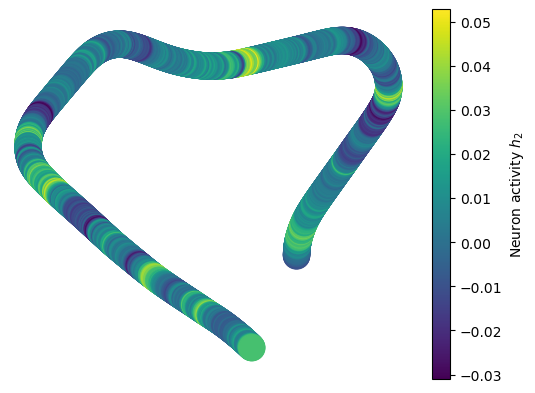}&
\includegraphics[width=.19\linewidth]{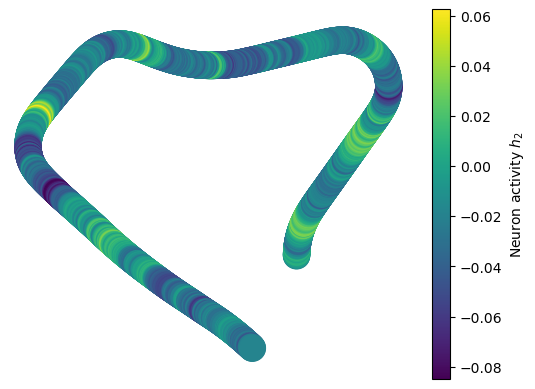}&
\includegraphics[width=.19\linewidth]{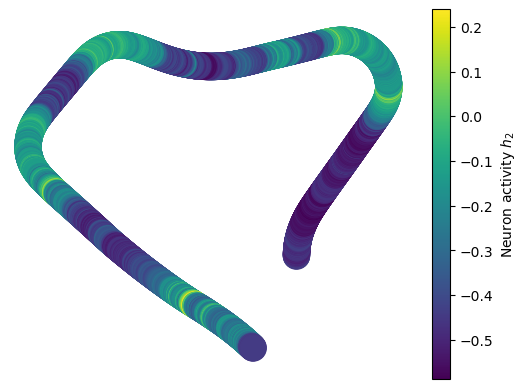}&
\includegraphics[width=.19\linewidth]{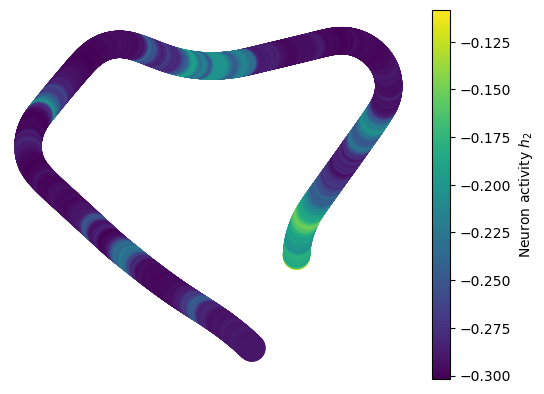}\\
%
%
%
\end{tabular}
\caption{Interpretability. The neural activity of two command cells in the learned policy for the Lane-Keeping Task, projected over time on the 1km road driven in summer.}
%
\label{fig:neural_activity}
\vspace*{-1ex}
\end{figure}

As another interpretability metric, we assess how the neural activity within a Lane-Keeping RNN-policy changes during deployment in the closed-loop simulation. Specifically, we are interested in identifying neurons exhibiting an activity matching the geometry of the trajectory. We conduct tests for a 1 km long drive, in both summer and winter. In Figure~\ref{fig:neural_activity}, we illustrate the activity of States $h_1$-$h_2$, corresponding to Cells 1-2 in the RNN policy, for summer, respectively. In the Appendix, we illustrate the activity of Cells 1-2 for winter. Our results reveal that the LSTM, GRU, MGU cells have scattered, hard-to-interpret activity, respectively. In contrast, LRCU-S cells demonstrate a very gentle varying activity, that aligns very well with the road's trajectory. This is most likely due to their double-liquid resistance and capacitance time constant.

\begin{table}[t]
    \centering
    \caption{Interpretability. Absolute correlation between the neural activity and the trajectory of the road. Values closer to 1 indicate stronger correlation. Averaged over 3 runs.}
    \label{tab:correlations}
   \begin{tabular}{lcc}
   \toprule
         &  Summer & Winter \\
    \midrule
       LSTM & $0.400 \pm 0.299$ & $0.450 \pm 0.262$ \\
       GRU & $0.363 \pm 0.256$ & $0.389 \pm 0.248$  \\
       MGU & $0.317 \pm 0.227$ & $0.308 \pm 0.232$ \\
       LRCU-A & $0.500 \pm 0.314$ & $0.533 \pm 0.292$\\
       LRCU-S & $\mathbf{0.645 \pm 0.321}$ & $\mathbf{0.766 \pm 0.243}$\\
    \bottomrule
   \end{tabular}
\end{table}

As shown in Table~\ref{tab:correlations}, we also compute the absolute-value cross-correlation metric, between the final prediction sequence of the policy (which aligns with the road's trajectory), and the sequence of activities of individual neurons. By considering the absolute value of the cross-correlations, we ensure that equal importance is given to both positive (that is excitatory) and negative (that is inhibitory) behaviors. This method yields values within the range of $[0, 1]$, where values close to zero indicate little to no correlation, and values close to one mean a very high correlation. The results presented in Table~\ref{tab:correlations} reinforce the observations from Figure \ref{fig:neural_activity}, which indicate that LRCU-based models, exhibit higher absolute-correlation values with the road's trajectory compared to the other models.

\section{Discussion, Scope and Conclusion}
\label{sec:conclusion}
We introduced Liquid-Resistance Liquid-Capacitance networks (LRCs), a Neural ODE model which is more accurate, more efficient, and biologically more plausible, compared to neuroscience's Electric Equivalent Circuits, and (saturated) Liquid Time-Constant networks (STCs and LTCs).  

Due to their gentle varying behavior LRCs are easy to integrate and applicable to a wide range of scenarios, including continuous processes and datasets sampled regularly or irregularly. LRCs thus represent a promising advancement in the field of bio-inspired deep learning. LRCs integrated with explicit Euler of order 1, and a time-step of 1 (LRCUs), have a very similar structure to gated RNNs. They not only demonstrate comparable performance to LSTMs, GRUs, and MGUs on several benchmarks, but also exhibit significantly higher interpretability, demonstrated in the autonomous Lane-Keeping task.

%

Despite the limitation of the benchmarks used, which we acknowledge could be expanded in subsequent work, our focus on analyzing the properties of our advanced bio-inspired model and its performance across diverse scenarios demonstrates its potential and lays a solid foundation for future exploration. Our results suggest that incorporating additional concepts from neuroscience can advance machine learning models, offering opportunities for future research to explore and further enhance model performance. 


\bibliographystyle{splncs04}
\bibliography{LRC_Festschrift/refs1}

\newpage
\appendix
\setcounter{theorem}{0}
\section{Appendix}

This appendix provides the proofs of the two theorems stated for LRCs, discusses the RNN form of LRCUs and their connection to GRUs and other gated recurrent units, and gives additional details for the experimental evaluation, demonstrating the accuracy, efficiency, and interpretability of LRCs.

\subsection{Proofs of the Theorems}
\label{app:theorems}

In this section we provide a detailed proof of the Theorems stated in this paper. The LRC Lipschitz Theorem first derives the local Lipschitz constant of LRCs and then shows that it is smaller than the one of STCs. The LRC Generalization Theorem, first computes the error of LRCs on the validation time series, and then shows that it is less than the one of the STCs, considering the validation set being drawn from in-distribution.

\subsubsection{Proof of the Lipschitz constant Theorem}\label{appendix:smoothness}

\begin{theorem}[LRC Lipschitz]
    Let the Lipschitz constant $\lambda^s$ bound the sensitivity of an STC-model instance of Equation~\eqref{eq:sltcs}, with respect to its inputs and hidden states. Then, there exists an associated LRC-model instance of Equation~\eqref{eq:lcrs}, with a Lipschitz constant $\lambda^r$, such that $\lambda^r<\lambda^s$.
\end{theorem}
%
\begin{proof} 
Let $T$ be the time horizon of the prediction, $K$ be the dimension of the output, $m$ the dimension of the states, $n$ the dimension of the input and $Q\in \mathbb{R}^{K\times m}$ define the linear output layer.
Let the output of the STC be defined by $o^s\,{=}\,Q\,h^s$, with $\dot{h}^s$ satisfying Equations~\eqref{eq:sltcs}, hidden states $h^s$, inputs $x$, and linear output $Q h^s$, and with trained synaptic parameters for $Q$ and $\dot h^s$. Let an associated LRC be defined by $o^r\,{=}\,Q\,h^r$, with $\dot{h}^r$ satisfying Equations~\eqref{eq:lcrs}. If we keep in the LRC all the parameters as they were trained for the STC, and only adapt the elastance $\epsilon(w)$, then it holds that $\dot h^r = \epsilon(w) \odot \dot h^s$.

The Lipschitz constant $\lambda$ of the output $o(y)$ computed by the LRC and STC networks, is defined such that $\|o(y_1)-o(y_2)\|_2\le \lambda \|y_1 - y_2\|_2$ for all $y_1,y_2$ in the input space of these networks. From the Mean Value Theorem it holds that for every $y_1, y_2$ it exists a $c$ such that $\|o(y_1)-o(y_2)\|_2\le \|\nabla_y o (c)\|_2 \|y_1 - y_2\|_2$. Thus, if a maximum exists, then $\max_c \|\nabla_y o (c)\|_2$ is a Lipschitz constant.

We now want to prove that there exists a choice of parameters for $\epsilon(w)$, such that the Lipschitz constant of the output $o^r$ of the LRC with respect to its hidden states and input $y=[h,x]$, is less than the one of $o^s$ in the STC. So the goal is to state conditions on the parameters of $\epsilon(w)$ such that:
\begin{equation}
    \max_y\|\nabla_y o^s\|_2 \le \lambda^s\ \wedge
    \max_y\|\nabla_y o^r\|_2 \le \lambda^r \quad \Rightarrow\quad \lambda^r < \lambda^s.
\end{equation}
Let us use $o, h$ when stating an equation which holds for STCs as well as LRCs:
%
\begin{align}
    o = Qh, \quad &\quad 
    o_k = \sum_{i=1}^m q_{ki}h_i,\\
    \pdv{o_k}{y_j} &= \sum_{i=1}^m q_{ki}\pdv{h_i}{y_j}.\label{eq:derivative o}
\end{align}
With $\Delta_t$ the time difference between two outputs, and by using Leibniz integral rule it holds that:
\begin{align}
    \left|\pdv{h_{t}}{y_{t,j}}\right|
    &= \left|\pdv{h_{t-\Delta t}}{y_{t,j}}  + \pdv{}{y_{t,j}}\int_{t-\Delta t}^{t}\dot h dt\right| \\
    &= \left|\underbrace{\pdv{h_{t-\Delta t}}{y_{t,j}}}_{\le 1}  + \int_{t-\Delta t}^{t}\pdv{\dot h}{y_{t,j}} dt \right|\\
    &\le 1 + \left|\int_{t-\Delta t}^{t}\max_y \pdv{\dot h}{y_{t,j}}dt\right|.
\end{align}
As $\dot h$ is time-independent, $\max_y\partial_{y_j}\dot h$ is the same for all $t$, thus:
\begin{align}
    \left|\pdv{h_{t}}{y_{t,j}} \right|&\le 1 + \left|\Delta t\cdot \max_y \pdv{\dot h}{y_{t,j}}\right|
\end{align}
\begin{align}
    \overset{\eqref{eq:derivative o}}{\Rightarrow} \left|\pdv{o_k}{y_j}\right| &\le \sum_{i=1}^m |q_{ki}|\left(1 + \Delta t\cdot\max_y\left|\pdv{\dot h_i}{y_j}\right|\right) = v_{kj}\label{eq:ub_output}\\
    \max_y\|\nabla_y o\|_2 &\le \|V\|_2 = \lambda.\label{eq:lambda}
\end{align}
Here $V\in\mathbb{R}^{K\times m}$ contains the values $v_{kj}$ from Equation~\eqref{eq:ub_output}.
%
When bounding $V$ later on in the proof, we will need the following known upper bounds of derivatives:
\begin{align}
    \sigma'(x) \le 0.25, &\quad
    \tau'(x) \le 1.\label{eq:derivatives activation functions}
\end{align}

The derivative of the sigmoid function $\sigma'(x) = \sigma(x)(1 - \sigma(x))$ has a known upper bound of 0.25, which occurs when x = 0. The derivative of the hyperbolic tangent function, $\tau'(x) = \mathrm{sech}^2(x)$, has a known upper bound of 1, which occurs when x = 0. Using these bounds, for the symmetric version of elastance in Equation~\eqref{eq:symmetrice} it holds that:
\begin{align}
    \epsilon'(w_i) = \sigma'(w_i + k_i) - \sigma'(w_i - k_i)  \\ \le \max\sigma'(w_i + k_i) - \min\sigma'(w_i - k_i) \le 0.25.
\end{align}
Further, we will need later on, that for $\epsilon(w_i)\in(0,1)$ it holds that:
\begin{align}
    \frac{\epsilon'(w_i)}{1-\epsilon(w_i)}\le\epsilon(w_i).\label{eq:epsilon_bound}
\end{align}
For asymmetric $\epsilon$ this is straightforward as the derivative of a sigmoid is  $\sigma'(x)=\sigma(x)\cdot (1-\sigma(x))$. For symmetric $\epsilon$ we need to reformulate the equations using that $\sigma(w_i+k_i)\ge\sigma(w_i-k_i)$ for $k_i\ge 0$:
\begin{align}
    \frac{\epsilon'(w_i)}{1-\epsilon(w_i)} = &
    \frac{\sigma(w_i+k_i)(1-\sigma(w_i+k_i))}{1-\sigma(w_i+k_i)+\sigma(w_i-k_i)}\\
    &-\frac{\sigma(w_i-k_i)(1-\sigma(w_i-k_i))}{1-\sigma(w_i+k_i)+\sigma(w_i-k_i)}\\
    &\le \frac{\sigma(w_i+k_i)(1-\sigma(w_i+k_i))}{1-\sigma(w_i+k_i)}\\
    &-\frac{\sigma(w_i-k_i)(1-\sigma(w_i+k_i))}{1-\sigma(w_i+k_i)} \\
    &= \epsilon(w_i).
\end{align}
Now we can derive an upper bound for the sensitivity of $\dot h^s_i$ to the input or hidden state $y_j$:
\begin{align}
    \left|\pdv{\dot h^s_i}{y_j}\right| &\le \left|\pdv{\sigma(f_i)}{y_j}h^s_i\right| +\underbrace{\left|\sigma(f_i)\pdv{h^s_i}{y_j}\right|}_{\le 1} + \left|\pdv{\tau(u_i)}{y_j}e_{li}\right|\\
    &\le |\sigma'(f_i)g_{ji}\sigma'(a_{ji}y_j + b_{ji})a_{ji}h^s_i|
    \\ & +|\tau'(u_i)k_{ji}\sigma'(a_{ji}y_j + b_{ji})a_{ji}e_{li}| +1\\
    &\le 0.0625 |g_{ji}a_{ji}h^s_i| + 0.25|k_{ji}a_{ji}e_{li}| +1  \\ &= \max_y \left|\pdv{\dot h^s_i}{y_j} \right|.\label{eq:upper bound partial h}
\end{align}
For the sensitivity of the LRC it holds that:
\begin{align}
    \dot{h^r_i} &= \epsilon(w_i) \cdot \dot{h^s_i}\\
    \left|\pdv{\epsilon(w_i)}{y_j}\right| &= |\epsilon'(w_i)\cdot o_{ji}|
    \\
    \left|\pdv{\dot{{h^r_i}}}{y_j}\right| &=
    \left|\epsilon(w_i)\cdot \pdv{\dot h^s_i}{y_j}  + \pdv{\epsilon(w_i)}{y_j}\cdot \dot{h}^s_i\right|\label{eq:derivative tilde}.
\end{align}
Here $o_{ji}$ are the trainable parameters of $w_i$.
As $\pdv{\epsilon(w_i)}{y_j} = \epsilon(w_i)(1-\epsilon(w_i))o_{ji}$, for $\epsilon(w_i)\in \{0,1\}$ it follows from Equation~\eqref{eq:derivative tilde} that $|\partial_{y_j}\dot{{h^r_i}}| = |\partial_{y_j} \dot h^s_i|$. For $\epsilon(w_i)\in (0,1)$:
\begin{align}
    \left|\pdv{\dot h^r_i}{y_j}\right| &=
    \left|\epsilon(w_i)\cdot \pdv{\dot h^s_i }{y_j} + \pdv{\epsilon(w_i)}{y_j}\cdot \dot{h}^s_i\right|\\
    &\le \epsilon(w_i)\cdot \left|\pdv{\dot h^s_i}{y_j}\right| + \epsilon'(w_i)\cdot |o_{ji}|\cdot |\dot{h}^s_i|\label{eq:upper bound partial h^r}.
\end{align}
Now let us choose the trainable parameters $o_{ji}$ in such a way, that:
\begin{align}
    |o_{ji}|\le \min\{0.0625 |g_{ji}a_{ji}|, 0.25|k_{ji}a_{ji}|\}= o^*_{ji}\label{eq:upper bound o}.
\end{align}
Then for this choice, and using Equation~\eqref{eq:epsilon_bound} it holds that:
\begin{align}
    \frac{\epsilon'(w_i)}{1-\epsilon(w_i)} |o_{ji}||\dot h^s_i|  \\
    \overset{\eqref{eq:epsilon_bound}}{\le} &\epsilon(w_i)|o_{ji}||\dot h^s_i|\\
    \overset{\eqref{eq:sltcs},\eqref{eq:derivatives activation functions}}{\le} &\epsilon(w_i)|o_{ji}|(|h^s_i| + |e_{li}|)\\
    \le &|o_{ji}||h^s_i| + |o_{ji}||e_{li}|\\
    \overset{\eqref{eq:upper bound o}}{\le} &0.0625 |g_{ji}a_{ji}h^s_i| + 0.25|k_{ji}a_{ji}e_{li}| + 1 \\
    \overset{\eqref{eq:upper bound partial h}}{=} &\max_y \left|\pdv{\dot h^s_i}{y_j}\right|,
\end{align}
which is equivalent to:
\begin{align}
    \epsilon'(w_i)\cdot |o_{ji}|\cdot |\dot{h}^s_i| \le (1-\epsilon(w_i)) &\max_y \left|\pdv{\dot h^s_i}{y_j}\right|\\
    &\Leftrightarrow\nonumber\\
    \epsilon(w_i)\cdot \max_y \left|\pdv{\dot h^s_i}{y_j}\right| + \epsilon'(w_i)\cdot |o_{ji}|\cdot |\dot{h}^s_i| \le &\max_y \left|\pdv{\dot h^s_i}{y_j}\right|\\
    &\overset{\eqref{eq:upper bound partial h^r}}{\Rightarrow} \nonumber\\
    \max_y\left|\pdv{\dot h^r_i}{y_j}\right| \le
    &\max_y \left|\pdv{\dot h^s_i}{y_j}\right|\label{eq:compare_pdv}.
\end{align}

If for all $i,j$ Equation~\eqref{eq:upper bound o} holds, and if there exists at least one pair $\bar i,\bar j$ such that $|o_{\bar j \bar i}| < o^*_{\bar j \bar i}$, then also Equation~\eqref{eq:compare_pdv} holds with inequality for these indices ($\max_y\partial_{y_{\bar j}}\dot h^r_{\bar i} < \max_y \partial_{y_{\bar j}}\dot h^s_{\bar i}$) and thus by using Equations~\eqref{eq:ub_output}-~\eqref{eq:lambda} it follows that:
\begin{align}
    v^r_{kj} &\le v^s_{kj}  \quad \forall j=\{1,\dots,m+n\}\\
    v^r_{k \bar j} &< v^s_{k \bar j}\\
    \Rightarrow \lambda^r &< \lambda^s.
\end{align}
This proves that the Lipschitz constant of LRCs is smaller than the Lipschitz constant of STCs.
\end{proof}

\subsubsection{Proof of the Generalization Theorem}\label{appendix:accuracy}

\begin{theorem}[LRC Generalization]
    Let $h^s$ be an STC-model instance of Equation~\eqref{eq:sltcs} with $h^r$ being an associated LRC-model instance of Equation~\eqref{eq:lcrs} with $\lambda^r < \lambda^s$ (Theorem.~\ref{thm:smoothness}). Let $o^T_t$ be training labels and $\hat o^{T,r}_t$  and $\hat o^{T,s}_t$ be the predictions of the models $h^r$ and $h^s$ respectively. Let both models have a small training loss $\|o^T_t-\hat{o}^{T}_t\|_2 < \epsilon_T$ with small $\epsilon_T>0$.
    Let $y^T_t$ and $y^V_t$ be respectively training and validation input.
    Let the validation set be similar to the training set (drawn in-distribution) in a sense that for every time $t$ it holds that $\|y^T_t-y^V_t\|_2<\epsilon_y$ and $\|o^T_t-o^V_t\|_2<\epsilon_o$ with small $\epsilon_y, \epsilon_o > 0$.
    Then it holds that the upper bound of the validation loss of the LRC-model is smaller than the one of the STC-model.
\end{theorem}

\begin{proof}
Let $o^T_1,\dots, o^T_T$ be the labels of the training set, $o^V_1,\dots, o^V_T$ be the ones of the validation set and $\hat o^{T}_t$ and $\hat o^{V}_t$ be respectively the output of the models of the training and the validation set. The STC or LRC model will be denoted in the superscript as $s$ or $r$. When not using one of these superscripts, the statement holds for both models. Let $T$ be the time horizon of the prediction.

Let the loss function for the STC and LRC networks be defined as usual, as follows:
\begin{align}
    L(o,\hat o) = \frac1{T}\sum_{t=1}^T \|o_t - \hat o_t\|_2.
\end{align}
Then for the validation loss it holds that:
\begin{align}
    L&(o^V,\hat o^{V}) = \frac1{T}\sum_{t=1}^T \|o^V_t - \hat o^V_t\|_2\\
    &= \frac1{T}\sum_{t=1}^T \|o^V_t-o^T_t + o^T_t-\hat o^T_t + \hat o^T_t - \hat o^V_t\|_2\\
    &\le \frac1{T}\sum_{t=1}^T \|o^V_t-o^T_t\|_2 + \|o^T_t-\hat o^T_t\|_2 + \|\hat o^T_t - \hat o^V_t\|_2\\
    &\le \frac1{T}\sum_{t=1}^T \epsilon_o + \epsilon_T + \|\hat o^T_t - \hat o^V_t\|_2. \label{eq:intermediate result validation loss}
\end{align}
As $\hat o^V_t$ is a function of the input $y^V_t$, we will switch to the notation: $\hat o^T_t = \hat o(y^V_t)$. Moreover, as we assumed that $\|y^T_t-y^V_t\|_2<\epsilon_y$, we can use a Taylor-series approximation:
\begin{align}
    \hat o(y^V_t) &\approx \hat o(y^T_t) + (y^V_t-y^T_t)^T\nabla_y \hat o(y^V_t)\\
    &\Leftrightarrow \nonumber \\
    \hat o(y^V_t) - \hat o(y^T_t) &\approx (y^V_t-y^T_t)^T\nabla_y \hat o(y^V_t) \\
    &\Rightarrow \nonumber \\
    \|\hat o(y^V_t) - \hat o(y^T_t)\|_2 &\approx \|(y^V_t-y^T_t)^T\nabla_y \hat o(y^V_t)\|_2 \\
    &\le \|(y^V_t-y^T_t)\|_2 \cdot \|\nabla_y \hat o(y^V_t)\|_2 \\
    &\le \epsilon_y \cdot \lambda.\label{eq:upper bound validation loss}
\end{align}
Here, $\lambda$ is the Lipschitz constant of the model. Putting this upper bound into Equation~\eqref{eq:intermediate result validation loss}, we can set the upper bound of the validation loss $\bar L (o^V,\hat o^V)$ as follows:
\begin{align}
    L(o^V,\hat o^V) \le \epsilon_o + \epsilon_T + \epsilon_y\cdot \lambda = \bar L (o^V,\hat o^V).
\end{align}
Finally, by using Theorem~\ref{thm:smoothness}, it holds that
\begin{align}
    \bar L (o^V,\hat o^{V,r}) = \epsilon_o + \epsilon_T + \epsilon_y\cdot \lambda^r \\ < \epsilon_o + \epsilon_T + \epsilon_y\cdot \lambda^s = \bar L (o^V,\hat o^{V,s}).
\end{align}
This proves that the validation loss of LRCs is smaller than the validation loss of STCs.
\end{proof}

\subsection{Liquid-Resistance Liquid-Capacitance Units (LRCU)}\label{appendix:lrcu_as_gatedrnn}
\label{app:lrcu}

In this section, we first introduce the RNN formulation of LRCUs, which as discussed in the paper, are a very accurate and efficient Euler integration of order one of LRCs, with a time step of one. Since LRCUs turn out to be a new form of gated recurrent units, we explore their relation to GRUs.

\subsubsection{LRCUs as an RNN} 
Starting from the Neural-ODEs model of saturated biological neurons with chemical synapses (the saturated EECs) of Equation~\eqref{eq:sltcs}, we have shown that first considering a liquid capacitance (elastance) as in Equation~\eqref{eq:lcrs} and then discretizing the ODEs leads to a very accurate and efficient gated RNN, which we called an LRCU. The formulation of this discrete unit is the following: 
\begin{equation}
\label{eq:lrcu}
\begin{array}{c}
    h_{i,t} =(1-\epsilon(w_{i,t})\,\sigma(f_{i,t}))\,h_{i,t-1} + \epsilon(w_{i,t})\,\tau(u_{i,t})\,e_{li}\\[2mm]
    f_{i,t} = \sum_{j=1}^{m+n} g_{ji}\sigma(a_{ji}y_{j,t}+b_{ji}) + g_{li}\\[2mm]
    u_{i,t} = \sum_{j=1}^{m+n} k_{ji}\sigma(a_{ji}y_{j,t}+b_{ji}) + g_{li}\\[2mm]
    w_{i,t} = \sum_{j=1}^{m+n} o_{ji}y_{j,t}+ p_{i}.
\end{array}
\end{equation}
Here, one can use either an asymmetric or a symmetric form of elastance $\epsilon$ in Equations~(\ref{eq:asymmetrice}-\ref{eq:symmetrice}), respectively. We will denote the first choice as LRCU-A, and the second choice as LRCU-S.

Next, we show how can one re-formulate GRUs and make connections to LRCUs. MGUs and LSTMs could be related to LRCUs, with minor modifications, in a similar manner.

\subsubsection{LRCUs versus GRUs}

In this section we explore the connections between LRCUs and GRUs. The general form of a gated recurrent units (GRU) is according to~\cite{cho2014learning}, an RNN of the following form:
\begin{equation}
\label{eq:gru}
\begin{array}{c}
h_{i,t} = (1-\sigma(f_{i,t}))\,h_{i,t-1} + \sigma(f_{i,t})\,\tau(u_{i,t})\\[3mm]

f_{i,t} = \sum_{j=1}^{m+n} a^{f}_{ji}\, y_{j,t} + b^{f}_{j}\\[2mm]
u_{i,t} = \sum_{j=1}^{m+n} a^{u}_{ji}\, y'_{j,t} + b^{u}_{j}\\[2mm]
r_{i,t} = \sum_{j=1}^{m+n} a^{r}_{ji}\, y_{j,t} + b^{r}_{j}.
\end{array}
\end{equation}
Here, vector $y_{t}$ occurring in functions $f_{i,t}$ and $r_{i,t}$, is defined as before, as $y_{t}\,{=}\,[h_{t-1},x_t]$. However, vector $y'_{t}$ occurring in $u_{i,t}$ is defined as $y'_{t}\,{=}\,[\sigma(r_t)\odot{}h_{t-1},x_t]$. 
Thus, previous state $h_{t-1}$ is pointwise scaled in the update part $\tau(u_{i,t})$ of the GRU, with a nonlinear state-and-input dependent function $\sigma(r_{i,t})$, whose parameters are to be learned. This function is called in GRUs a Reset Gate (RG). Moreover, the state-and-input dependent function $\sigma(f_{i,t})$ is called in GRUs an Update Gate (UG). 

The RG determines how previous state $h_{t-1}$ is used in the update $\tau(u_{i,t})$. The UG $\sigma(f_{i,t})$ controls the amount $(1-\sigma(f_{i,t}))$ of the previous state $h_{i,t-1}$, to be remembered in the next state. However, this UG also controls the amount of the update $\tau(u_{i,t})$ to be considered in the next state, by using it to multiply the update. One can identify the GRU's UG with the LRCU's smoothen (elastance) gate.

Given the above discussion, GRUs can also be understood as the ordinary difference equations associated with the Neural ODEs below. From this, one can get back to the original form of Equations~\eqref{eq:gru}, using an explicit Euler integration scheme of order one with a unit time difference:
\begin{equation}
\label{eq:gru_ode}
\begin{array}{c}
\dot{h}_{i} = \sigma(f_{i})(-h_{i} + \tau(u_{i}))\\[2mm]
f_{i} = \sum_{j=1}^{m+n} a^{f}_{ji}\, y_{j} + b^{f}_{j}\\[2mm]
u_{i} = \sum_{j=1}^{m+n} a^{u}_{ji}\, y'_{j} + b^{u}_{j}\\[2mm]
r_{i} = \sum_{j=1}^{m+n} a^{r}_{ji}\, y_{j} + b^{r}_{j}.
\end{array}
\end{equation}
Here the vector $y\,{=}\,[h,x]$ is defined as before, and vector $y'\,{=}\,[\sigma(r)\odot{}h,x]$. The RG occurring in $y'$ determines how state $h$ is used in the update part $\tau(u_{i})$ of the ODE.
The LRC's time constant (TC) thus consists of a liquid-resistance liquid-capacitance, while the GRU's TC consists of a liquid capacitance, only. As a consequence, GRUs have a less expressive TC but more expressive update.

\subsection{Experiments for LRCs and LRCUs}\label{appendix:experiment_details}

In this section, we provide additional implementation details about our experimental evaluation of LRC in wide range of ODE modeling tasks and LRCUs on the popular time-series benchmarks for gated recurrent units. The benchmarks considered were the Localization Data for Person Activity, the IMDB Movie Sentiment Classification, the Permuted Sequential MNIST, and a relatively complex autonoumous-driving in the Lane-Keeping Task.

In general, LRC(U)s have more trainable parameters per cell. Therefore, we are training the other models with more cells to ensure a fair comparison between them. For the experiments, we determined the learning rate using the logarithmic scale of $\{10^{-4}, 10^{-3}, 10^{-2}\}$ by the validation loss. For the individual tests, see the more detailed setups below. They are run on Ubuntu 22.04 with Nvidia Tesla T4.

\subsubsection{Neural ODE Experiments}

For ODE modeling tasks, we used a 3-layer (units of 32, 32 and 2) Neural ODE and the LRC (units of 16) with additional input and output mapping, which hold 1k trainable parameters. The hyperparameters are presented in Table~\ref{tab:neural_ode_hyperparam}.  We use sequences of 16 points during training and test the models on the whole sequences of 1000 data points giving them only the initial state information of $(x_0,y_0)$. As some ODE modeling tasks are more challenging, we allocated longer training iteration time for them. The training time is 0.4-0.5 seconds per iteration for both models.
\begin{table}[h]
\centering
\caption{Hyperparameters of the Neural ODE experiments.}
\begin{tabular}{ll}
\toprule
Variable                         & Value                
\\ \midrule
Learning rate                    & $10^{-3}$ \\
Batch size                       & $16$                   \\
Training sequence length         & $16$                   \\
Iterations                           & $1000/2000/4000$    \\
\bottomrule         
\end{tabular}
\label{tab:neural_ode_hyperparam}
\end{table}

\begin{itemize}
    \item Periodic Sinusoidal: $dx/dt = x \cdot (1- \sqrt{x^2+y^2}) -y$, $dy/dt = x+y \cdot (1-\sqrt{x^2+y^2})$.
    \item Spiral: $dx/dt = Ax$, where $A=[[-0.1, 3], [-3, -0.1]$.
    \item Duffing Oscillation: $ dx/dt = y$, $dy/dt = x - x^3$.
    \item Periodic Lotka-Volterra: $dx/dt = a \cdot x-b\cdot x \cdot y$, $dy/dt = -c\cdot y+d \cdot x \cdot y$ with $a=1.5$, $b=1$, $c=3$ and $d=1$.
    \item Asymptotic Lotka-Volterra:  $dx/dt = x \cdot (1-x)-x\cdot y$, $dy/dt = - y+d \cdot x \cdot y$ with $d=2$.
    \item Non-linear Lotka-Volterra:  $dx/dt = x \cdot (1-x)-a \cdot x\cdot y$, $dy/dt = y \cdot (1- y) + x \cdot y$ with $a=0.33$.
\end{itemize}

These examples originate from the TensorFlow implementation of~\cite{Chen2018NeuralOD}, tfdiffeq.

\subsubsection{Localization Data for Person Activity}
Considering that this is an irregularly sampled dataset, including separate extra timestep information per input, traditionally used gated units need further modification when dealing with this task. 

\begin{table}[h]
    \centering
    \caption{Number of trainable parameters in the Localization Data for Person Activity dataset. To have at least the same number of trainable parameters, we employed 64 cells for LRCUs and 100 cells in the other models.}
    \begin{tabular}{cccccc}
        \toprule
        & LSTM & GRU & MGU & LRCU-A & LRCU-S \\
        \midrule
        Par. & 40k & 30k & 20k & 20k & 20k\\ 
        \bottomrule        
    \end{tabular}
    \label{tab:person_activity_param}
\end{table}
\begin{table}[h]
\centering
\caption{Hyperparameters of the Localization Data for Person Activity experiment.}
\vspace*{2mm}
\begin{tabular}{ll}
\toprule
Variable                         & Value                
\\ \midrule
Learning rate                    & $10^{-3}$ \\
Batch size                       & $128$                   \\
Training sequence length         & $32$                   \\
Epochs                           & $100$    \\
\bottomrule         
\end{tabular}
\label{tab:person_act_hyperparam}
\end{table}

For LSTMs, GRUs, and MGUs we concatenate the time-step information (difference in time between to inputs) directly with the input features. This time step is used as the $\Delta_t$ value in LRCUs. LSTM, GRU and MGU networks contain 100 units, while LRCUs have 64 units.

In Table~3 we provide the total number of trainable parameters used by each model. Moreover, in Table~4 we provide information about the hyperparameters used by all models to solve this task.

\subsubsection{IMDB Movie Sentiment Classification}
In the IMDB review dataset we keep the 20,000 most frequent words and truncate the sequences up to 256 characters. Token embeddings of size 64 are used. LSTM, GRU and MGU have 100 units, while LRCU variants have 64 cells in the networks.

\begin{table}[h]
    \centering
    \caption{Total number of trainable parameters in the IMDB Movie Sentiment Classification task. We used 64 cells in the LRCUs and 100 cells in the other models.}
    \vspace*{2mm}
    \begin{tabular}{cccccc}
        \toprule
        & LSTM  & GRU & MGU & LRCU-A & LRCU-S\\
        \midrule
        Par. & 70k & 50k & 35k & 40k & 40k\\
        \bottomrule
    \end{tabular}
    \label{tab:imdb_param}
\end{table}
\begin{table}[h]
\centering
\caption{Hyperparameters of the IMDB Movie Sentiment Classification task.}
\vspace*{2mm}
\begin{tabular}{ll}
\toprule
Variable                         & Value                
\\ \midrule
Learning rate                    & $10^{-3}$ \\
Batch size                       & $64$                   \\
Training sequence length         & $256$                   \\
Epochs                           & $30$    \\
\bottomrule         
\end{tabular}
\label{tab:imdb_hyperparam}
\end{table}

In Table~\ref{tab:imdb_param} we provide the total number of trainable parameters used by each model to solve this task. Moreover, in Table~\ref{tab:imdb_hyperparam} we provide information about the hyperparameters used to solve this task.

\begin{table}[h]
    \centering
    \caption{Total number of parameters in the Permuted Sequential MNIST task. LRCUs use 64 neurons, whereas the other 100 neurons, each. All gated-RNN models were run for 3 seeds.}
    \vspace*{2mm}
    \begin{tabular}{cccccc}
        \toprule
        & LSTM & GRU & MGU & LRCU-A & LRCU-S\\
        \midrule
        Par. & 40k & 30k & 20k & 20k & 20k\\
        \bottomrule
    \end{tabular}
    \label{tab:psmnist_param}
\end{table}

\subsubsection{Permuted Sequential MNIST}
As in the other classification tasks, traditionally used gated networks, such as LSTMs, GRUs and MGUs, have 100 units, and the proposed LRCUs contain 64 units. In Table~7 we provide details about the hyperparameters used in this task. Moreover, in Table~8, we provide the total number of trainable parameters for each model used.

\begin{table}[h]
\centering
\caption{Hyperparameters of the Permuted Sequential MNIST experiment.}
\begin{tabular}{ll}
\toprule
Variable                         & Value                
\\ \midrule
Learning rate                    & $10^{-3}$ \\
Batch size                       & $64$                   \\
Training sequence length         & $784$                   \\
Epochs                           & $200$    \\
\bottomrule         
\end{tabular}
\label{tab:psmnist_hyperparam}
\end{table}

During the training of this task, the computed loss values of MGUs became NaNs in 2 out of the 3 runs in the middle of the experiments. 

In Table~\ref{tab:psmnist_param} we provide the total number of trainable parameters used by each model to solve this task. Moreover, in Table~\ref{tab:psmnist_hyperparam} we provide information about the hyperparameters used to solve this task.

\subsubsection{Lane-Keeping Task}
In this section we provide details about the Lane-Keeping-Task RNN-policy architecture. In Table~\ref{tab:conv-head} below we describe the convolutional head used by the policy. 

\begin{table}[h!]
\centering
\caption{The shape and size of the layers in the convolutional head of the RNN-policy used to solve the Lane-Keeping Task. Settings are adapted from~\cite{farsang2024learning}.}
\vspace*{2mm}
\begin{tabular}{ll}
\toprule
Layer Type  & Settings \\ 
\midrule
Input                   & Input shape: (48, 160, 3)\\
Image-Norm.   & Mean: 0, Variance: 1\\
Conv2D                  & Filters: 24, Kernel size: 5, Stride: 2\\
Conv2D                  & Filters: 36, Kernel size: 5, Stride: 1\\
MaxPool2D               & Pool size: 2, Stride: 2  \\
Conv2D                  & Filters: 48, Kernel size: 3, Stride: 1\\
MaxPool2D               & Pool size: 2, Stride: 2\\
Conv2D                  & Filters: 64, Kernel size: 3, Stride: 1\\
MaxPool2D               & Pool size: 2, Stride: 2\\
Conv2D                  & Filters: 64, Kernel size: 3, Stride: 1 \\
Flatten                 & -\\
Dense                   & Units: 64\\
\bottomrule         
\end{tabular}
\label{tab:conv-head}
\end{table}

For the Lane-Keeping task, the total number of parameters of the recurrent part is 8k, where the number of neurons used for LSTMs, GRUs, MGUs, LRCU-A, and LRCU-S are 23, 28, 38, 19, and 19 respectively. This choice ensures a similar number of trained parameters.

In Table~\ref{tab:hyperparam} we provide a description of the hyperparameters used in the Lane-Keeping Experiment. They had the same value for all the RNN models compared.

\begin{table}[h!]
\centering
\caption{Hyperparameters of the Lane-Keeping experiment, adapted from \cite{farsang2024learning}.}
\begin{tabular}{ll}
\toprule
Variable                         & Value                \\ \midrule
Learning rate                    & cosine decay, $5 \cdot 10^{-4}$ \\
Optimizer                        & AdamW \\ 
Weight decay                     & $10^{-6} $              \\
Batch size                       & $32$                   \\
Training sequence length         & $32$                   \\
Epochs                           & $100$    \\   
\bottomrule         
\end{tabular}
\label{tab:hyperparam}
\end{table}

\begin{table}[]
    \centering
    \caption{Training time per epoch.}
    
    \begin{tabular}{lr}
      \toprule
        Model   &  Time/epoch\\
      \midrule
      LSTM & 2.57 min\\
      GRU & 2.50 min\\
      MGU & 3.08 min\\
      LTC    & 10.06 min\\
      STC    & 10.13 min\\
      LRCU-A  & 4.05 min\\
      LRCU-S  & 4.16 min\\
      \bottomrule
    \end{tabular}
    \label{tab:training_time_lanekeeping}
\end{table}

\subsection{Interpretability Experiments for LRCUs}\label{appendix:interpretability}
In this section we provide additional details for the interpretability experiments for the Lane-Keeping Task.

\begin{figure}[t]
\settoheight{\tempdima}{\includegraphics[width=.08\linewidth]{example-image-a}}%
\centering\begin{tabular}{@{ }c@{ }c@{ }c@{ }c@{ }c@{ }c@{ }c@{ }}
 & & \small{LSTM} & \small{GRU} &  \small{MGU} & \small{LRCU-A} & \small{LRCU-S} \\
\rowname{0.0}&
\includegraphics[width=.145\linewidth]{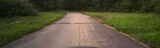}&
\includegraphics[width=.145\linewidth, decodearray={0.25 1 0.25 1 0.25 1}]{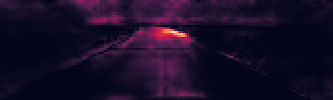}&
\includegraphics[width=.145\linewidth, decodearray={0.25 1 0.25 1 0.25 1}]{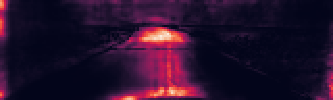}&
\includegraphics[width=.145\linewidth, decodearray={0.25 1 0.25 1 0.25 1}]{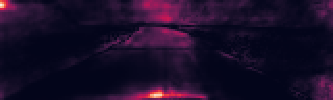}&
\includegraphics[width=.145\linewidth, decodearray={0.25 1 0.25 1 0.25 1}]{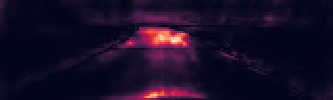}&
\includegraphics[width=.145\linewidth, decodearray={0.25 1 0.25 1 0.25 1}]{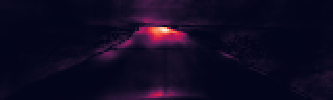}\\
\rowname{0.1}&
\includegraphics[width=.145\linewidth]{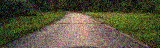}&
\includegraphics[width=.145\linewidth, decodearray={0.25 1 0.25 1 0.25 1}]{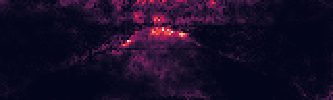}&
\includegraphics[width=.145\linewidth, decodearray={0.25 1 0.25 1 0.25 1}]{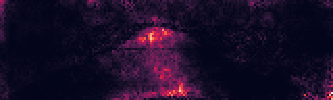}&
\includegraphics[width=.145\linewidth, decodearray={0.25 1 0.25 1 0.25 1}]{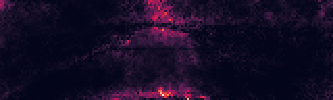}&
\includegraphics[width=.145\linewidth, decodearray={0.25 1 0.25 1 0.25 1}]{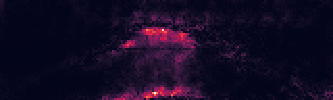}&
\includegraphics[width=.145\linewidth, decodearray={0.25 1 0.25 1 0.25 1}]{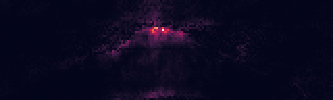}\\
\rowname{0.2}&
\includegraphics[width=.145\linewidth]{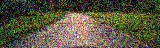}&
\includegraphics[width=.145\linewidth, decodearray={0.25 1 0.25 1 0.25 1}]{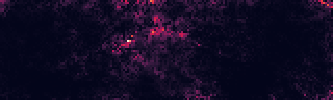}&
\includegraphics[width=.145\linewidth, decodearray={0.25 1 0.25 1 0.25 1}]{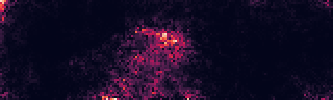}&
\includegraphics[width=.145\linewidth, decodearray={0.25 1 0.25 1 0.25 1}]{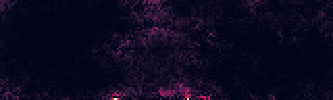}&
\includegraphics[width=.145\linewidth, decodearray={0.25 1 0.25 1 0.25 1}]{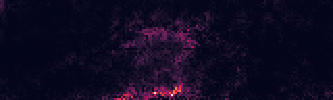}&
\includegraphics[width=.145\linewidth, decodearray={0.25 1 0.25 1 0.25 1}]{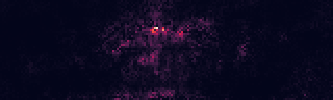}\\
\end{tabular}
\caption{Interpretability. Attention in summer. Column~1 shows the input of the network: Row~1, without additional noise, Rows~2-3, with Gaussian noise of variance $\sigma^2=0.1$ and $\sigma^2=0.2$, respectively. Remaining columns correspond to the networks, showing their attention to the same input image. One can observe how much the focused areas get distorted in the presence of noise.}%
\vspace{4ex}
\label{fig:saliency_summer}
\end{figure}

\begin{figure}[h]
\settoheight{\tempdima}{\includegraphics[width=.05\linewidth]{example-image-a}}%
\centering\begin{tabular}{@{ }c@{ }c@{ }c@{ }c@{ }c@{ }c@{ }c@{ }}
 & & LSTM & GRU &  MGU & LRCU-A & LRCU-S \\
\rowname{0.0}&
\includegraphics[width=.145\linewidth]{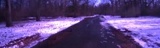}&
\includegraphics[width=.145\linewidth, decodearray={0.25 1 0.25 1 0.25 1}]{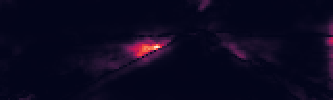}&
\includegraphics[width=.145\linewidth, decodearray={0.25 1 0.25 1 0.25 1}]{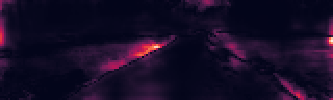}&
\includegraphics[width=.145\linewidth, decodearray={0.25 1 0.25 1 0.25 1}]{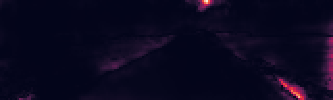}&
\includegraphics[width=.145\linewidth, decodearray={0.25 1 0.25 1 0.25 1}]{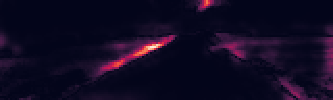}&
\includegraphics[width=.145\linewidth, decodearray={0.25 1 0.25 1 0.25 1}]{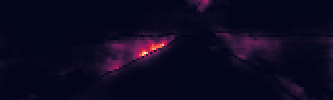}\\
\rowname{0.1}&
\includegraphics[width=.145\linewidth]{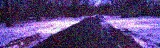}&
\includegraphics[width=.145\linewidth, decodearray={0.25 1 0.25 1 0.25 1}]{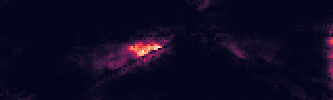}&
\includegraphics[width=.145\linewidth, decodearray={0.25 1 0.25 1 0.25 1}]{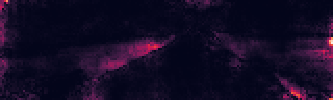}&
\includegraphics[width=.145\linewidth, decodearray={0.25 1 0.25 1 0.25 1}]{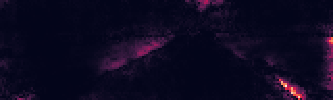}&
\includegraphics[width=.145\linewidth, decodearray={0.25 1 0.25 1 0.25 1}]{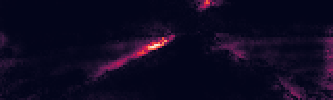}&
\includegraphics[width=.145\linewidth, decodearray={0.25 1 0.25 1 0.25 1}]{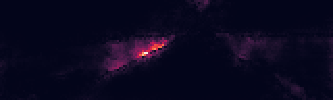}\\
\rowname{0.2}&
\includegraphics[width=.145\linewidth]{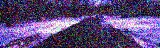}&
\includegraphics[width=.145\linewidth, decodearray={0.25 1 0.25 1 0.25 1}]{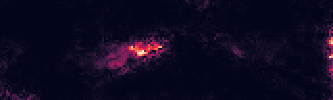}&
\includegraphics[width=.145\linewidth, decodearray={0.25 1 0.25 1 0.25 1}]{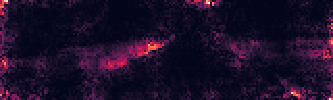}&
\includegraphics[width=.145\linewidth, decodearray={0.25 1 0.25 1 0.25 1}]{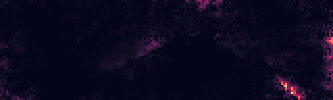} &
\includegraphics[width=.145\linewidth, decodearray={0.25 1 0.25 1 0.25 1}]{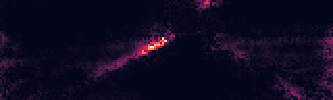}&
\includegraphics[width=.145\linewidth, decodearray={0.25 1 0.25 1 0.25 1}]{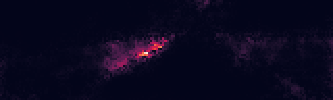}\\
\end{tabular}
\caption{Attention maps in the winter season. In general, the focus shifts from the road to the side of the road compared to summer. Unimportant regions on the off-road are attended by LSTM, GRU and MGU.  A winter input image and its saliency maps of the analyzed models are displayed, with increasing noise of Gaussian noise of $\sigma^2=0.1$ and $\sigma^2=0.2$ variances, in the second and third rows. }%
\label{fig:saliency_winter}
\end{figure}

Figure \ref{fig:saliency_summer} shows our results in the summer season, where lighter-highlighted regions indicate the attention of the network. We found that the LSTM takes into account irrelevant parts of the image, during its decision-making. The rest of the models have most of the attention on the road, and LRCU networks especially, focus on the horizon. In Figure~\ref{fig:saliency_winter}, we provide the attention maps for the winter season. 

By injecting a zero mean Gaussian noise with a variance of 0.1 and 0.2 into the tests, respectively, we can measure how robust their attention is. The noisy attention maps are displayed in the second (variance 0.1) and third (variance 0.2) rows of Figure~\ref{fig:saliency_winter}. This noise was not present during training, 

By measuring the change of attention, we also compute the Structural Similarity Index (SSIM)~\cite{SSIM} of the images pairwise, between the noise-free and the noisy saliency map for each model, respectively. This technique assesses the similarity between two images, with a focus on image degradation. An SSIM value of one denotes full similarity, whereas one of zero, denotes zero similarity. In our particular case, where we are comparing attention-map images, values close to one mean that the extra noise leads to less distortion in the attention map, which is a very desirable aim in the development of robust controllers. The SSIM value for $1600{-}1600$ pairs of images, used in the attention-map comparisons between noise-free and noisy, summer and winter images, respectively, are presented in Figure~\ref{fig:ssim}. All results in this table are averaged over three runs for each model. As one can see, LRCUs demonstrate much more robust attention in the presence of Gaussian noise.

\begin{figure}[h]
\settoheight{\tempdima}{\includegraphics[width=.2\linewidth]{example-image-a}}%
\centering\begin{tabular}{@{}c@{ }c@{ }c@{ }c@{}c@{}c@{}}
& LSTM & GRU & MGU & LRCU-A & LRCU-S \\
%
%
\rowname{Cell 1}&
\includegraphics[width=.17\linewidth]{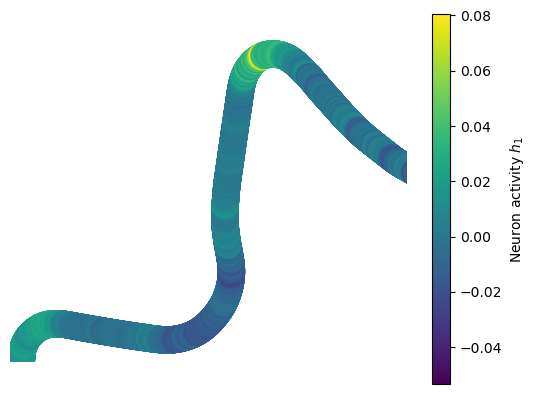}&
\includegraphics[width=.17\linewidth]{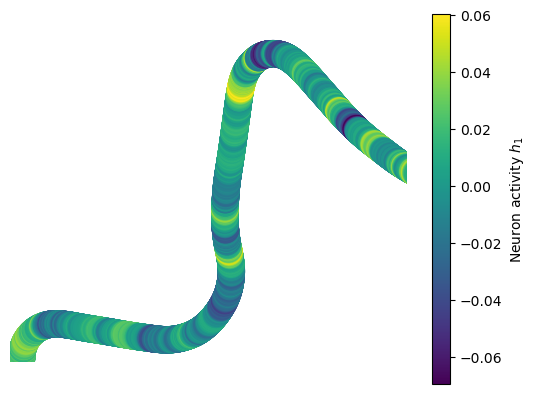}&
\includegraphics[width=.17\linewidth]{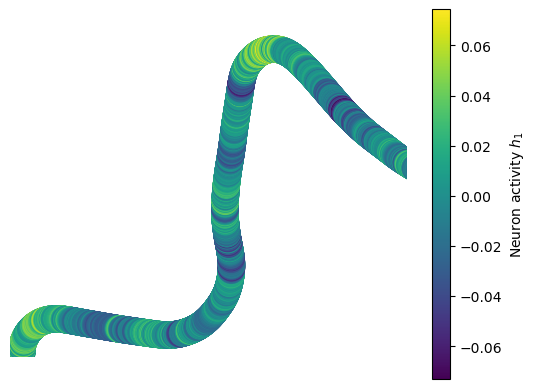}&
\includegraphics[width=.17\linewidth]{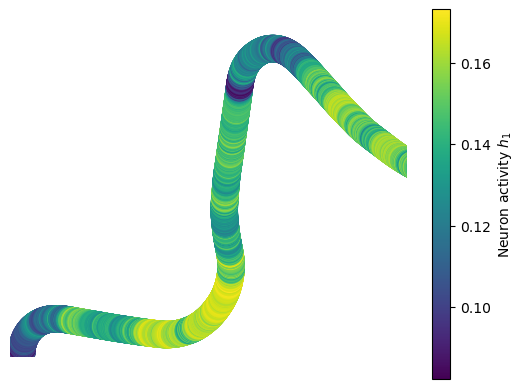}&
\includegraphics[width=.17\linewidth]{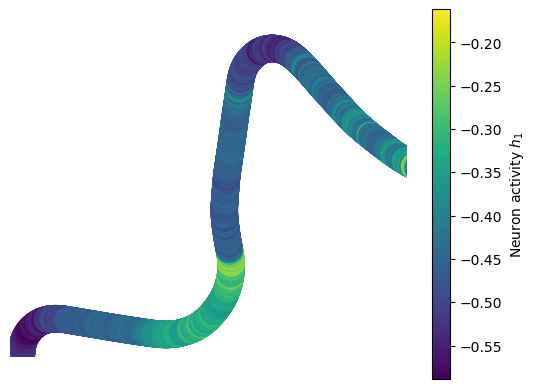}\\
\rowname{Cell 2}&
\includegraphics[width=.17\linewidth]{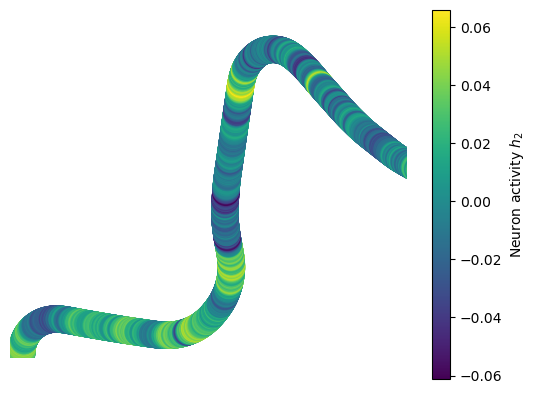}&
\includegraphics[width=.17\linewidth]{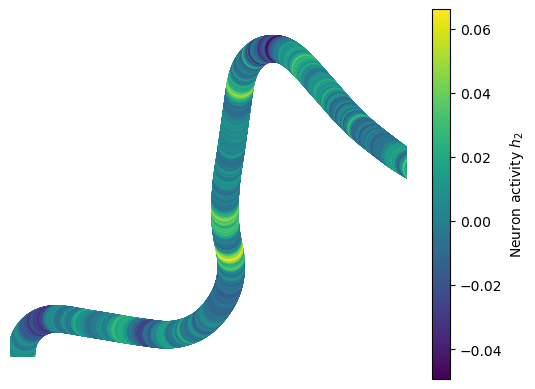}&
\includegraphics[width=.17\linewidth]{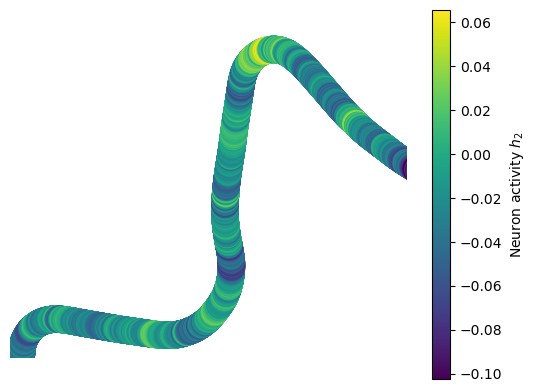}&
\includegraphics[width=.17\linewidth]{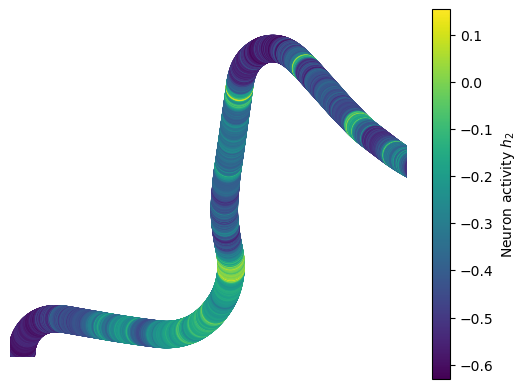}&
\includegraphics[width=.17\linewidth]{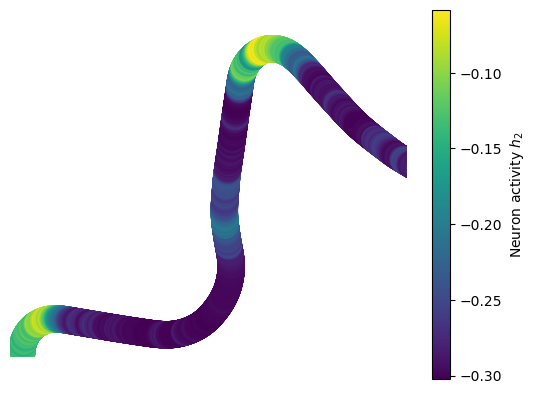}\\
%
\end{tabular}
\caption{Interpretability. The neural activity of two cells in the learned policy for the Lane-Keeping Task, projected over time on the 1km road driven in winter, for three popular gated recurrent units (LSTMs, GRUs, and MGUs) and for LRC units (LRCU-A, and LRCU-S). It is very hard to visually match the neural activity of LSTM, GRU, and MGU cells to the traversed road geometry,  respectively. However, LRCU-S cells especially, display a very gentle varying and identifiable pattern during driving: Cell 1 is responsible for turning right, and Cell 2 for turning left. }
%
\label{fig:neural_activity_winter}
\end{figure}

In Figure~\ref{fig:neural_activity_winter} we provide the neural activity of two cells of the lane-keeping policy for the winter season. As one can clearly see, the LRCU-S has a very gentle varying behavior closely matching the road geometry: Cell 1 is responsible for turning left, and Cell 2 is responsible for turning right.

\end{document}